\documentclass[final,5p,times,authoryear]{elsarticle}

\usepackage{graphicx}
\usepackage[tight]{subfigure}
\usepackage{amsmath,amssymb}
\usepackage{adjustbox}
\usepackage{tikz}
\usepackage{pgfplots}
\usepackage{colortbl}
\usepackage{arydshln}
\usepackage{algorithm,algpseudocode}

\journal{ISPRS Journal of Photogrammetry and Remote Sensing}

\graphicspath{}
\DeclareGraphicsExtensions{.pdf,.jpg,.png}

\def\Ext#1{\ifnum #1=1{Probability-Level-Fusion}\else{\ifnum#1=2{Logit-Level-Fusion}\else{\ifnum#1=3{Feature-Level-Fusion}\else{Pixel-Level-Fusion}\fi}\fi}\fi}
\def\ext#1{\Ext{#1}}

\usepackage[normalem]{ulem}
\usepackage{soul}
\definecolor{revcolor}{rgb}{0.0, 0.0, 1.0}

\def\rev#1{{#1}} 

\def\mypar#1{\addvspace{.1cm}\noindent{\bf #1}} 
\def\mypardot#1{\mypar{#1.}} 
\newcommand\hfilll{\hspace{0pt plus 1filll}} 

\begin{document}

\begin{frontmatter}

\title{Weakly Supervised Instance Attention for Multisource Fine-Grained Object
    Recognition with an Application to Tree Species Classification}
\author[bilkent]{Bulut Aygunes}
\ead{bulut.aygunes@bilkent.edu.tr}
\author[metu]{Ramazan Gokberk Cinbis}
\ead{gcinbis@ceng.metu.edu.tr}
\author[bilkent]{Selim Aksoy\corref{cor}}
\ead{saksoy@cs.bilkent.edu.tr}
\cortext[cor]{Corresponding author.
    Tel: +90 312 2903405; fax: +90 312 2664047.}
\address[bilkent]{Department of Computer Engineering, Bilkent University,
    Ankara, 06800, Turkey}
\address[metu]{Department of Computer Engineering, Middle East Technical
    University, Ankara, 06800, Turkey}

\begin{abstract}
Multisource image analysis that leverages complementary spectral, spatial,
and structural information benefits fine-grained
object recognition that aims to classify an object into one of many similar
subcategories. However, for multisource tasks that involve relatively small
objects, even the smallest registration errors can introduce high uncertainty
in the classification process. We approach this problem from a weakly supervised
learning perspective in which the input images correspond to larger neighborhoods
around the expected object locations where an object with a given class label
is present in the neighborhood without any knowledge of its exact location.
The proposed method uses a single-source deep instance attention model with parallel
branches for joint localization and classification of objects, and extends
this model into a multisource setting where a reference source that is assumed
to have no location uncertainty is used to aid the fusion of multiple sources
in four different levels: probability level, logit level, feature level,
and pixel level. We show that all levels of fusion provide higher accuracies
compared to the state-of-the-art, with the best performing method of feature-level
fusion resulting in $53\%$ accuracy for the recognition of $40$ different types
of trees, corresponding to an improvement of $5.7\%$ over the best performing
baseline when RGB, multispectral, and LiDAR data are used. We also provide an
in-depth comparison by evaluating each model at various parameter complexity
settings, where the increased model capacity results in a
further improvement of $6.3\%$ over the default capacity setting.
\end{abstract}


\end{frontmatter}

\section{Introduction}
\label{sec:Introduction}

Advancements in sensors used for remote sensing enabled spectrally rich images to be acquired at very high spatial resolution. Fine-grained object recognition, which aims the classification of an object as one of many similar subcategories, is a difficult problem manifested by these improvements in sensor technology \citep{Oliveau:2017, Branson:2018, Sumbul:2018}. The difficulty of distinguishing subcategories due to low variance between classes is one of the main characteristics of this problem that differs from traditional object recognition and classification tasks studied in the remote sensing literature. Other distinguishing features of fine-grained object recognition are the difficulty of collecting samples for a large number of similar categories, which can cause the training set sizes to be very limited for some classes, and the class imbalance that makes the traditional supervised learning approaches to overfit to the classes with more samples. This makes it necessary to develop new methods for fine-grained classification that could cover the shortfalls of the traditional object recognition methods regarding these problems.

One way to help decrease the confusion inherent to the data in fine-grained classification is to gather complementary information by utilizing multiple data sources, which can provide more distinguishing properties of the object of interest. For example, a high-resolution RGB image can give details about texture,
color, and coarse shape, whereas a multispectral (MS) image can provide richer spectral content and LiDAR data can yield information about the object height. However, the question of how to combine the data from multiple sources does not have a straightforward answer. Therefore, it is an open research problem to find a method to benefit from the distinct contents of the sources as effectively as possible.

The common assumption of most multisource image analysis methods is that the data sources are georeferenced or co-registered without any notable errors that may prevent the pixel- or feature-level fusion of the sources. This can be a valid assumption for tasks like land cover classification in which the classes of interest (e.g., water, forest, impervious surfaces) are significantly larger compared to the registration errors \citep{CHEN201727}. However, for multisource tasks that involve the classification of relatively small objects such as trees---similar to the problem we focus on in this paper---even the smallest registration errors can introduce high uncertainty among different sources and between the sources and the ground truth labels. Furthermore, it is not always possible to mitigate this uncertainty by trying to discover pixel-level correspondences between the sources due to the fine-grained nature of the problem and for other reasons such as differences in the imaging conditions, viewing geometry, topographic effects, and geometric distortions \citep{Han:2016}.

\begin{figure}
\centering
\includegraphics[width=\linewidth]{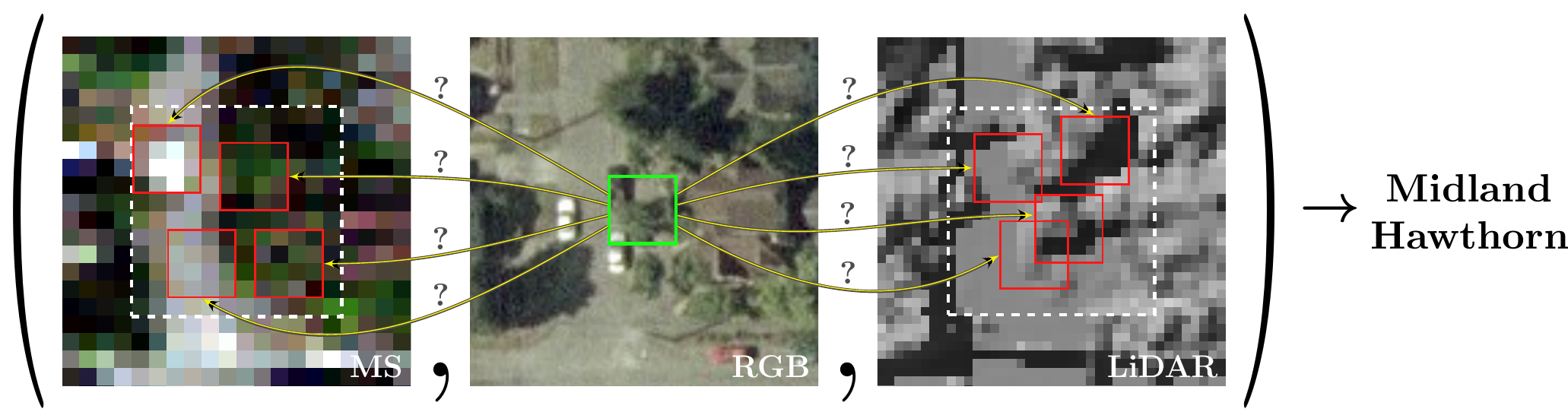}
\caption{Illustration of our multisource fine-grained object recognition problem. The sources are only approximately registered, therefore, it is unclear which pixels in the low-resolution sources (MS and LiDAR) correspond to the object of interest centered in the high-resolution reference source (RGB). Our goal is to implicitly tackle the registration uncertainties through instance attention to correctly predict the object class using information from all sources.}
\label{fig:Scenario}
\end{figure}

The fine-grained recognition problem studied in this paper involves the classification of street trees using RGB, MS, and LiDAR data. Although the very high-resolution RGB images are manually inspected with respect to the reference tree locations in the GIS-based ground truth, there is still high location uncertainty in the MS and LiDAR data that contain trees of $4 \times 4$ and $8 \times 8$ pixels, respectively, due to the aforementioned reasons. To cope with this uncertainty introduced by registration errors and small sizes of the target objects, we crop tiles larger than the object sizes around the reference locations given by the point GIS data to ensure that each tree falls inside its corresponding tile. With such images covering larger neighborhoods than the typical object size, the problem becomes a weakly supervised learning (WSL) problem in the sense that the label of each image provides information about the category of the object it contains, but does not yield any information regarding its location in the neighborhood. The problem is illustrated in Figure~\ref{fig:Scenario}.

To the best of our knowledge, the only related work that studied the multisource fine-grained recognition problem is that of \citet{Sumbul:2019}, where an attention mechanism over a set of candidate regions was used with guidance by a reference source to obtain a more effective representation of the sources which are fused together for the final classification. However, in that scheme, the attention mechanism simply aims to maximize the discriminative power of the attention-driven representation. While that approach yields promising results, it is susceptible to overfitting to \mbox{\em accidental} correlations appearing in training examples, and may learn to put too much emphasis on background features. In this work, we instead utilize a stronger WSL-based formulation that aims to induce an \mbox{\em instance attention} behavior: instead of estimating pooling weights of candidate regions, we estimate the relevance of each candidate region as a function of its spatial {\em and} semantic distinctiveness. Here, therefore, we aim to incorporate the prior knowledge that in most cases one (or very few) local regions actually belong to the object of interest in a local neighborhood.

The method proposed in this paper loosely builds upon our preliminary work \citep{Aygunes:2019}, which has shown that weakly supervised learning objective can be repurposed to improve single-source object recognition when the images contain high location uncertainty.
In this paper, as our main contribution, we extend this idea to the multisource setting with a number of novel information fusion schemes.
We first propose a more generalized version of our WSL-based instance attention model for single-source classification, which can be applied to any source with location uncertainty. Then, the proposed fusion schemes combine multiple additional sources that are processed with this instance attention method and a reference source that is assumed to have no uncertainty and is processed in a fully supervised fashion. Each proposed scheme aims to leverage the reference source to aid the instance attention branches by combining the reference with the additional sources in four different levels: probability level, logit level, feature level, and pixel level. We show that it is possible to benefit from the reference source with all levels of fusion, as they surpass the state-of-the-art baselines. As another contribution, we also propose a methodology to compare different models in a more principled way, by evaluating each model at various parameter complexity settings. The results of this experiment highlight the importance of investigating approaches at various model capacities to make fair comparisons, as comparing different methods each of which having a different single model capacity setting can be misleading. Overall, our results indicate that we obtain significant improvements over the state-of-the-art.

In the rest of the paper, we first present a summary of related work in Section~\ref{sec:RelatedWork} and give information about the data set we use in our experiments in Section~\ref{sec:DataSet}. We then describe our proposed methods in Section~\ref{sec:Methodology}. Next, we give details about our experimental setup, and present quantitative comparisons with several baselines and qualitative results in Section~\ref{sec:Experiments}. Finally, we provide our conclusions in Section~\ref{sec:Conclusions}.

\section{Related work}
\label{sec:RelatedWork}

\mypardot{Multisource image analysis}
There are many studies in the remote sensing literature that focus on multisource image analysis \citep{Chova:2015, DallaMura-PIEEE15}, which has also received the attention of data fusion contests \citep{Debes2014, Liao:2015, Taberner:2016, Yokoya:2018}. The research includes statistical learning methods such as dependence trees \citep{Datcu:2002}, kernel-based methods \citep{Valls:2008}, copula-based multivariate model \citep{Voisin:2014}, and active learning \citep{Zhang:2015}. Another well-studied problem is manifold alignment \citep{Tuia:2014, Hong-ISPRS19, Gao:2019} where the goal is to transfer knowledge learned in the source domain to a target domain. The underlying reason that necessitates this transfer is typically the spectral mismatch between the domains. In this paper, the main problem in the multisource analysis is the spatial mismatch among the image sources.

More recently, deep learning-based methods have focused on classification with pixel-level or feature-level fusion of multiple sources. Pixel-level fusion includes concatenation of hyperspectral and LiDAR data preprocessed to the same resolution, followed by a convolutional neural network (CNN) \citep{Morchhale2016}, while in feature-level fusion, hand-crafted \citep{Ghamisi:2017} or CNN-based \citep{Pibre:2017, Hu:2017, Xu:2018, IENCO201911} features, obtained from different data sources such as multispectral, hyperspectral, LiDAR, or SAR, are processed with convolutional and/or fully-connected layers to obtain the final decision.

\mypardot{Weakly supervised remote sensing}
WSL approaches in remote sensing have utilized class activation maps for object localization. For example, \citet{Ji:2019} combined the per-class activation maps from different layers of a convolutional network trained with image-level labels to obtain class attention maps and localize the objects. \citet{Wang:2020} proposed a modification to the U-Net architecture to enable using image-level weak labels corresponding to the majority vote of the pixel labels instead of pixel-level strong labels during the training for a binary segmentation task. \citet{Xu:2019} localized objects by using a combination of two different convolutional layers. \citet{Zhang:2019} suggested using gradients of network layers to obtain saliency maps for background and foreground classes. Similarly, \citet{Ma:2020} obtained saliency maps by utilizing gradients with respect to the input pixels to localize residential areas in aerial images.
\citet{ALI2020115} studied destruction detection from only image-level labels where each image was represented using a weighted combination of the patch-level representations obtained from a convolutional network. The weights were learned by using a single fully-connected layer that was trained using a sparsity loss.
\citet{li2020_cloud} introduced a global convolutional pooling layer to build a cloud detection network that was trained using block-level labels indicating only the presence or absence of clouds within image blocks without pixel-level annotations.
However, all of these approaches focus on a binary classification scenario.
For the multi-class setting, \citet{LI2018182} used pairs of images with the same scene-level labels to train a Siamese-like network for learning convolutional weights, and updated this network with a global pooling operation and a fully-connected layer to learn class-specific activation weights.
\citet{HUA2019188} used a linear combination of all channels in the output of a CNN-based feature extractor network to learn class-specific feature representations that were further combined in a recurrent neural network to learn class dependencies.
However, both of these approaches use a global combination of the convolutional channels where a single fully-connected layer is expected to learn the class attention mechanism. Furthermore, none of the approaches above considers the label uncertainty problem in a multisource setting.
In our case, while we do not aim to explicitly localize objects as in WSL studies, we propose a number of WSL-based formulations for addressing the spatial ambiguity in multisource object recognition.

Another important problem is the noise in the type of the label where a scene
or an object is labeled as another class instead of the true one. For example,
\citet{li2020_error_tolerant} proposed an error-tolerant deep learning approach
for remote sensing image scene classification by iteratively training a set of
CNN models that collectively partitioned the input data into a strong data set
with all models agreeing on the original labels and a weak data set with the
models producing different predictions. Each iteration built new CNN models
that used the union of the strong data set with the original labels and the
weak data set with the predicted labels as the new training data.
Here, we assume that the label itself is correct but an uncertainty exists in
its spatial location.

In a more relevant problem caused by misalignment of GIS maps and images used for building extraction, \citet{Zhang:2020} added a layer to a segmentation network to model the noise in the labels, and trained the model by calculating the loss using the noisy predictions and the noisy reference labels. Although such an approach can be useful in a task like building extraction, it might not be applicable for problems consisting of small objects like trees where a segmentation-based approach is not feasible due to the size and fine-grained nature of the objects.

\mypardot{Tree classification}
In this paper, we illustrate the proposed weakly supervised instance attention model and the multisource fusion schemes using a fine-grained street tree classification problem.
Novel applications involving street trees include \citep{Branson:2018} where aerial images and street-view panoramas were jointly used for fine-grained classification. The feature representations computed by deep networks independently trained for the aerial and ground views were concatenated and fed to a linear SVM for classification of 40 tree species. More recently, \citet{LAUMER2020125} improved the existing street tree inventories where the individual trees were referenced by only street addresses with accurate geographic coordinates that were estimated from multi-view detections in street-view panoramas.
The methods proposed in this paper are not specific to tree detection. Thus, a full review on tree species mapping is beyond the scope of this paper. We refer the reader to \citet{Fassnacht2016} that provides a review of such methods in which multispectral, hyperspectral, and LiDAR data sources have been the most widely used modalities.

\section{Data set}
\label{sec:DataSet}

\begin{figure}[t]
\centering
\includegraphics[width=\linewidth]{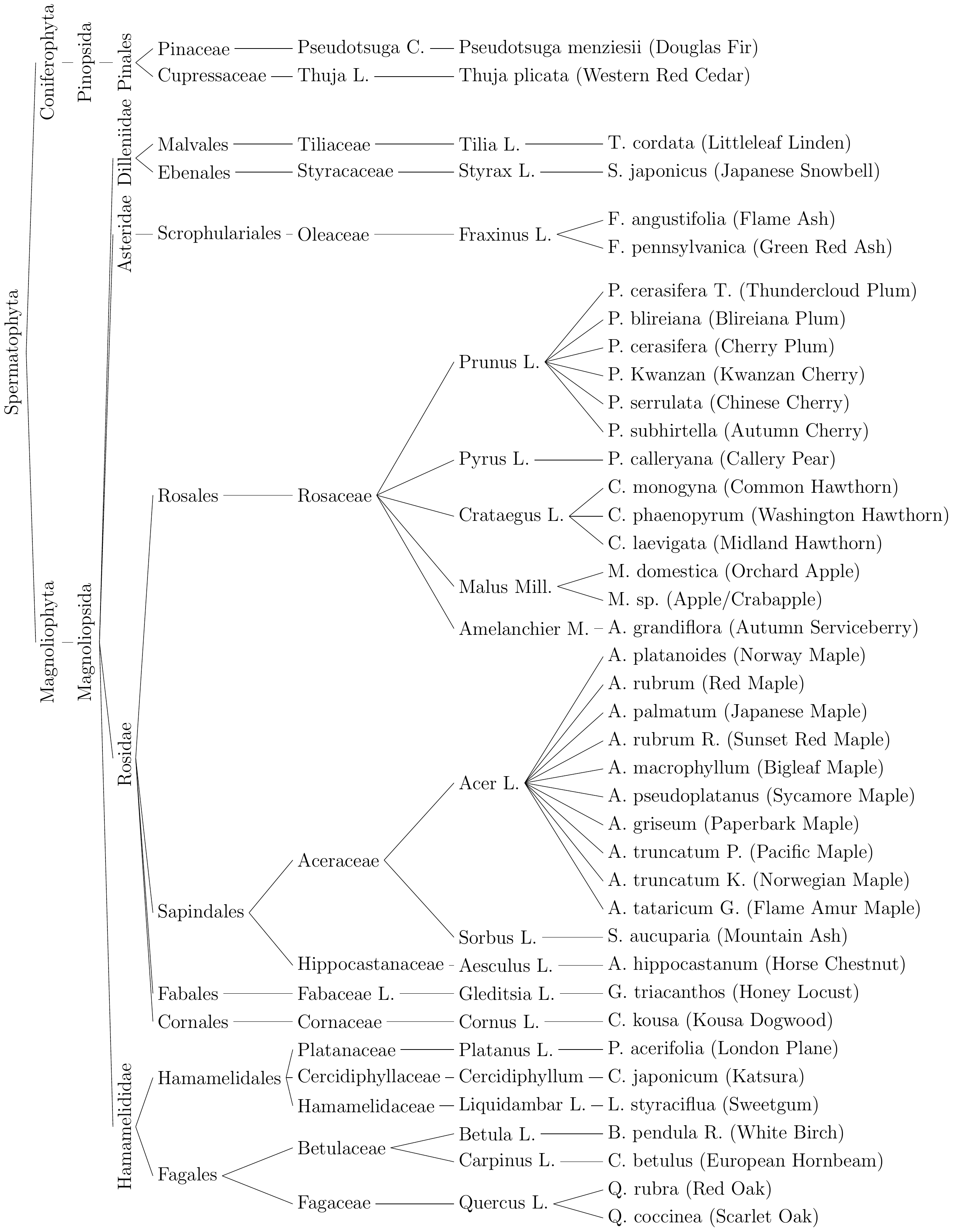}
\caption{Scientific classification of tree species. The taxonomy starts with
the Spermatophyta superdivision and continues with the names of division, class,
subclass, order, family, genus, and species in order. At the last level, the
common names are given in parentheses next to the scientific names.
The classifications are taken from \citet{tree_hierarchy}.}
\label{fig:Taxonomy}
\end{figure}

We conduct our experiments on the same data set as \citep{Sumbul:2019}, which is,
to our knowledge, the only multisource data set that includes a fine-grained set
of classes with an additional challenge of location uncertainty among the data
sources due to the sizes of the objects of interest. The data set consists of a
total of $48,\!063$ instances of street trees belonging to $40$ different classes.
The fine-grained nature of the data set is illustrated in Figure \ref{fig:Taxonomy}
where the scientific classification of tree species is presented as a hierarchy
in which most species differ only in the lowest levels. The number of samples
for each class in this highly imbalanced data set is shown in Table \ref{tb:Dataset}.

\begin{table}[t]
\centering
\caption{Class names and number of samples in each class in the data set.
The classes follow the same order as in Figure \ref{fig:Taxonomy}.}
\label{tb:Dataset}
\setlength{\tabcolsep}{2pt}
\begin{minipage}{0.48\linewidth}
\begin{adjustbox}{width=\linewidth}
\begin{tabular}{p{3.2cm}r}
\hline
Class name & Samples\\
\hline
Douglas Fir & 620\\
Western Red Cedar & 720\\
Littleleaf Linden & 1,626\\
Japanese Snowbell & 460\\
Flame Ash & 679\\
Green Red Ash & 660\\
Thundercloud Plum & 2,430\\
Blireiana Plum & 2,464\\
Cherry Plum & 2,510\\
Kwanzan Cherry & 2,398\\
Chinese Cherry & 1,531\\
Autumn Cherry & 621\\
Callery Pear & 892\\
Common Hawthorn & 809\\
Washington Hawthorn & 503\\
Midland Hawthorn & 3,154\\
Orchard Apple & 583\\
Apple/Crabapple & 1,624\\
Autumn Serviceberry & 552\\
Norway Maple & 2,970\\
\hline
\end{tabular}
\end{adjustbox}
\end{minipage}%
\hfill%
\begin{minipage}{0.48\linewidth}
\begin{adjustbox}{width=\linewidth}
\begin{tabular}{p{3.2cm}r}
\hline
Class name & Samples\\
\hline
Red Maple & 2,790\\
Japanese Maple & 1,196\\
Sunset Red Maple & 1,086\\
Bigleaf Maple & 885\\
Sycamore Maple & 742\\
Paperbark Maple & 467\\
Pacific Maple & 716\\
Norwegian Maple & 372\\
Flame Amur Maple & 242\\
Mountain Ash & 672\\
Horse Chestnut & 818\\
Honey Locust & 875\\
Kousa Dogwood & 642\\
London Plane & 1,477\\
Katsura & 383\\
Sweetgum & 2,435\\
White Birch & 1,796\\
European Hornbeam & 745\\
Red Oak & 1,429\\
Scarlet Oak & 489\\
\hline
\end{tabular}
\end{adjustbox}
\end{minipage}
\end{table}

For each tree sample, there are three images obtained from different sources:
an aerial RGB image with $1$ foot spatial resolution, an $8$-band WorldView-2
MS image with $2$ meter spatial resolution, and a LiDAR-based digital
surface model with $3$ foot spatial resolution.
The label and location information for the tree samples were obtained from the
point GIS data provided by the Seattle Department of Transportation in Washington
State, USA \citep{trees_data_set}. Since the GIS data set was constructed as
part of a carefully planned field campaign for inventory management, we assumed
that the class label for each tree is correct. However, we used visual interpretation
on the aerial RGB image, that corresponds to the data source with the highest
spatial resolution, as an additional effort to validate the consistency of the
tree locations \citep{Sumbul:2018}. Even though it was not possible to visually
confirm the tree category from the remotely sensed data, we made sure that the
provided coordinate actually coincided with a tree for every single one of the
samples. During this process, some samples had to be removed due to mismatches
with the aerial data, probably because of temporal differences between ground
data collection and aerial data acquisition.

For the confirmed $48,\!063$ samples, RGB images that were centered at the
locations provided in the point GIS data were cropped at a size of $25 \times 25$
pixels to cover the largest tree in the data set. The validation process for
the tree locations, together with the fact that RGB images have higher spatial
resolution than MS and LiDAR data, make RGB images suitable to be used as the
reference source. This choice can be made with a high confidence if there is
additional information regarding the georeferencing process of the data sources
and their compatibility with the ground truth object locations.

A tree in a $25 \times 25$ pixel RGB image corresponds to $4 \times 4$ pixels
in MS and $8 \times 8$ pixels in LiDAR data. Although each source was
previously georeferenced, registration errors can cause significant uncertainties
in the locations of small objects such as trees, especially in sources with lower
resolution such as MS and LiDAR. To account for the location uncertainties, we
use images that cover a larger neighborhood than a single tree. Specifically,
we use $12 \times 12$ pixel neighborhoods for MS and $24 \times 24$ pixel
neighborhoods for LiDAR \citep{Sumbul:2019}.

\def\zc{\phi^\mathit{cls}}
\def\zd{\phi^\mathit{loc}}
\def\sigc#1{[\sigma^\mathit{cls}(#1)]}
\def\sigd#1{[\sigma^\mathit{loc}(#1)]}
\def\wsl{\phi^\mathit{WSL}}
\def\enc{\phi^\mathit{enc}}
\def\prop{\phi^\mathit{prop}}
\def\reg{\phi^\mathit{region}}
\def\pred{\phi^\mathit{pred}}
\def\comb{\phi^\mathit{comb}}
\def\cnn{\phi^\mathit{CNN}_\mathit{ref}}
\def\feat{\omega}
\def\allx{x_{1:M}}

\begin{figure*}
\centering
\includegraphics[width=\linewidth,trim=0 0 27px 5px,clip]{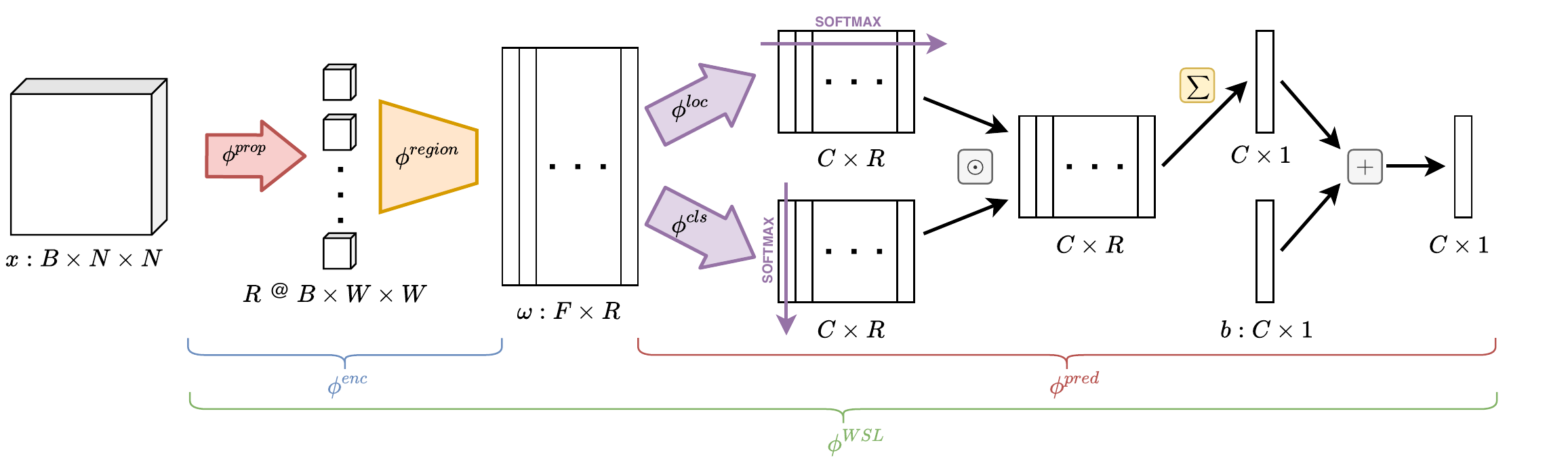}
\vspace{-1.5\baselineskip}
\caption{Illustration of the single-source weakly supervised instance attention model. Source index $m$ is omitted from all variables for simplicity. For the $m\textsuperscript{th}$ source, the model takes $x_m$ with size $B_m \times N_m \times N_m$ as the input. Region proposals are extracted using $\prop$ with a sliding window of size $W_m \times W_m$, resulting in $R_m = (N_m - W_m + 1)^2$ candidate regions. Each region is processed by $\reg$, which consists of 3 convolutional and 1 fully-connected layers. For MS, all convolutional layers have $64$ kernels of size $3 \times 3$ and no pooling is used afterwards. For LiDAR, $64$ convolution kernels with size $5 \times 5$ for the first two layers and $3 \times 3$ for the last layer are used, with max-pooling with kernel size $2 \times 2$ and stride $2$ after each convolutional layer. The fully-connected layer of $\reg$ outputs features of size $128$. A fully-connected layer ($\zd$) and softmax across regions are applied to the resulting matrix of per-region feature vectors $\feat^m$ for the localization branch, and another fully-connected layer ($\zc$) followed by softmax across classes is applied  for the classification branch. Hadamard product of the resultant matrices is taken and the result is summed over the regions. Finally, a bias vector $b^m$ is added to obtain the logits. ReLU activation is used throughout the network.}
\label{fig:wsddn}
\end{figure*}

\section{Methodology}
\label{sec:Methodology}
In this section, we first outline the weakly supervised multisource object recognition problem.
Then, we explain the single-source weakly supervised instance attention approach.
Finally, we present our four formulations to tackle the multisource recognition problem via instance attention.

\subsection{Weakly supervised multisource object recognition}
The multisource object recognition problem aims to classify an object into one of the $C$ classes by utilizing the images of the object coming from $M$ different sources. This corresponds to learning a classification function that takes the images $x_1, \ldots, x_M$ of an object from $M$ imaging sources
and outputs a class prediction $\hat{y} \in \{1, \ldots, C\}$. 

To cope with the location uncertainty in the data, we use images that cover a larger area than the objects of interest as the input to the model. More precisely, we assume that each image from the $m$\textsuperscript{th} source covers an $N_m \times N_m$ pixel neighborhood and contains a smaller object of size $W_m \times W_m$ with an unknown location. In such a setting, the ground truth for an image becomes a weak label in the sense that it does not hold any positional information about the object, which makes the problem a weakly supervised learning problem. 

In this work, we focus on RGB, MS, and LiDAR data which are acquired in different conditions (resolution, viewpoint, elevation, time of day, etc.). As a result, different registration uncertainties are present among the data sources, which cause the locations of the same object in the images from different sources to be independent of each other. This becomes one of the major challenges of the weakly supervised multisource object recognition problem.

\subsection{Single-source weakly supervised instance attention}

Location uncertainty in a WSL problem necessitates either explicit or implicit localization of the object to be classified. Successful localization of the object helps to obtain a more reliable representation by eliminating the background clutter, which in turn can improve the classification results. Following this intuition,
we construct our instance attention approach by adapting the learning formulation of Weakly Supervised Deep Detection Network (WSDDN) \citep{Bilen:2016}. WSDDN extracts $R$ candidate regions, some of which are likely to contain the object of interest, from an image $x$ using a region proposal operator, $\prop$. Each of these regions is transformed into a feature vector of size $F$ using a region encoder network, $\reg$, consisting of three convolutional layers and one fully-connected layer. We refer the reader to the caption in Figure \ref{fig:wsddn} for source-specific architectural details of the region encoder. For input $x \in X$,
\begin{equation}
\label{eq:feat}
    \enc: X \rightarrow \Omega
\end{equation}
collectively represents candidate region extraction ($\prop$) and region encoding ($\reg$) operations. Here, the resulting $\feat \in \Omega$ is an $F \times R$ matrix of per-region feature vectors. To simplify the notation, we define the remaining model components as a function of $\feat \in \Omega$.

After the region encoding operation, a localization branch scores candidate regions among themselves using softmax separately for each class, and outputs region localization scores:
\begin{equation}
    \sigd{\feat}_{ci}=\dfrac{\exp\!\big([\zd(\feat)]_{ci}\big)}{\sum_{r=1}^{R} \exp\!\big([\zd(\feat)]_{cr}\big)}
\end{equation}
where $[\zd(\feat)]_{ci}$ is the raw score of the $i$\textsuperscript{th} candidate region for the class $c$, obtained by the linear transformation $\zd$. Similarly, a parallel classification branch assigns region classification scores corresponding to the distribution of class predictions for each region independently:
\begin{equation}
    \sigc{\feat}_{ci}=\dfrac{\exp\!\big([\zc(\feat)]_{ci}\big)}{\sum_{k=1}^{C} \exp\!\big([\zc(\feat)]_{ki}\big)}
\end{equation}
where $[\zc(\feat)]_{ci}$ is the raw score obtained by the linear transformation $\zc$.

A candidate region that successfully localizes the true object is expected to yield both a higher localization score for the true object class than the other candidates and a higher classification score for the true class than the other classes. This property naturally yields a more instance-centric attention mechanism, compared to mainstream attention formulations that learn to weight candidate regions purely based on the discriminative power of the final attention-driven representation, see e.g.~\citet{Sumbul:2019}.  To implement this idea in a differentiable way, region localization and classification scores are element-wise multiplied and summed over all regions to obtain per-class localization scores on the image level:
\begin{equation}
\label{eq:pred}
    [\pred(\feat)]_c = \sum_{i=1}^{R} \sigd{\feat}_{ci} \odot \sigc{\feat}_{ci}.
\end{equation}

The formulation up to this point is quite generic. To implement it in an efficient and effective way for our weakly supervised fine-grained classification task, we use the following three ideas. 
First, we obtain candidate regions from the input image in a sliding window fashion with a fixed window of size $W_m \times W_m$, where $W_m$ is experimentally chosen for each source $m$. Second, we put an additional softmax layer at the end of the network. This additional softmax layer effectively incorporates the prior  knowledge that a local image area is likely to contain only a single class of interest. Finally, we add learnable per-class bias parameters to the class detection scores before the softmax operation, and update \eqref{eq:pred} as follows:
\begin{equation}
\label{eq:pred2}
    [\pred(\feat)]_c = \sum_{i=1}^{R} \sigd{\feat}_{ci} \odot \sigc{\feat}_{ci} + b_c
\end{equation}
where $b_c$ is the bias parameter for class $c$. Combining \eqref{eq:feat} and \eqref{eq:pred2}, we define the whole model as:
\begin{equation}
    \wsl(x) = \pred\!\big(\enc\!(x)\big),
\end{equation}
and class probabilities as:
\begin{equation}
\label{eq:outputwsl}
    P(c|x)=\big[\sigma\big(\wsl\!(x)\big)\big]_c
\end{equation}
where $P(c|x)$ is the probability predicted for the $c$\textsuperscript{th} class given image $x$ and $\sigma$ denotes the softmax operation.

\begin{figure*}
\centering
\includegraphics[width=\linewidth,trim=0 10px 7px 6px,clip]{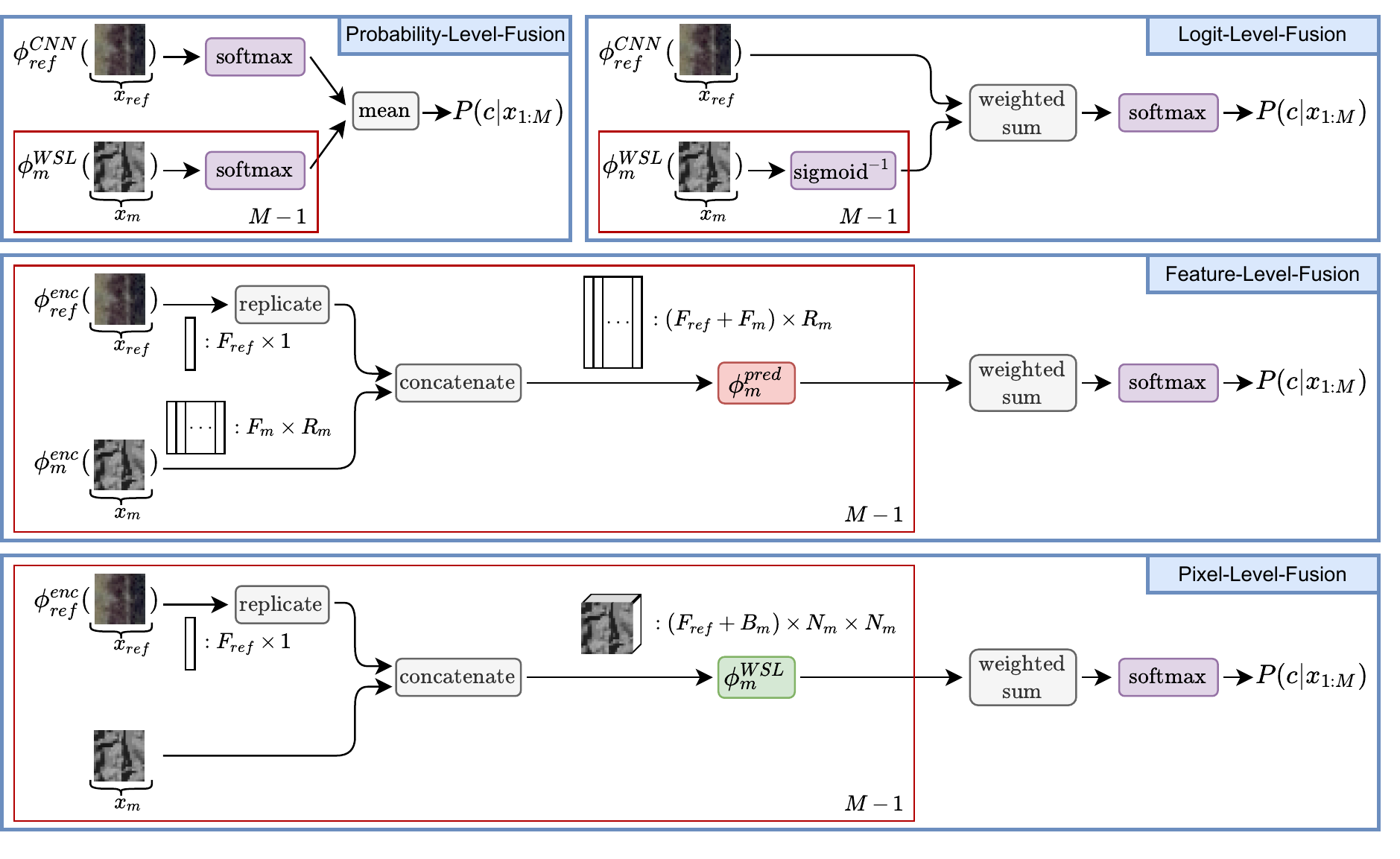}
\caption{Illustration of the proposed multisource WSL models. The reference source $x_1$ is represented with subscript $\mathit{ref}$, while subscript $m \in \{2 \ldots M\}$ is used for the additional sources. Plate notation is used to represent the repetitions in the model. Variables are described in the text.}
\label{fig:extensions}
\end{figure*}

\subsection{Multisource WSL models}

We base our models on the assumption that (at least) one of the $m$ sources does not have a high uncertainty
regarding the object location like the other sources. For simplicity, we refer to this source as $x_1$. This typically
corresponds to the high-resolution RGB imagery where georeferencing can be done relatively precisely and the object is
located centrally within the image. We aim to use this {\em reference source} to mitigate the uncertainty in the other
sources ($x_2,\ldots,x_M$), which are referred to as the {\em additional sources}. The ultimate goal is to increase the
overall classification performance by extracting (more) precise information from the additional sources.

To handle this ambiguity in weakly labeled sources, we propose four weakly supervised multisource models with instance attention. These
models handle the weakly supervised fusion problem progressively in different
levels, as indicated by their names:
(i) \Ext{1}, (ii) \Ext{2}, (iii) \Ext{3}, and (iv) \Ext{4}. In the following, we define and discuss these model schemes in detail.
Figure \ref{fig:extensions} illustrates the proposed models.

\mypardot{\Ext{1}}
In this model, we propose to combine additional data sources with the reference source by taking an average of the output probabilities of all sources:
\begin{equation}
    P(c|\allx) = \dfrac{1}{M}\sum_{m=1}^{M} P(c|x_m)
\end{equation}
where $P(c|x_m)$ is obtained as in \eqref{eq:outputwsl} using a separate instance attention network ($\wsl_m$) for each $m \in \{2,\ldots,M\}$ and a simple CNN ($\cnn$) for the reference source $x_1$. The only difference from \eqref{eq:outputwsl} is that the logits coming from the additional sources $\wsl_m(x_m)$ are divided by a temperature parameter $T_m < 1$ before the softmax operation to sharpen the output distribution, which is much smoother compared to the output of the reference $P(c|x_1)$.
\rev{A summary of this approach is given in Algorithm~\ref{alg:ext1}.}

Combining the sources at the probability level corresponds to giving equal weights to the outputs of all sources and allowing them to contribute to the final classification evenly. This could cause a source with a more confident prediction to have a higher impact on the final decision, which can be desirable or undesirable depending on the reliability of that particular source. The temperature parameter enables the model to pay more/less attention to some of the sources by adjusting the confidence levels of their predictions.

\begin{algorithm}[t]
\caption{\Ext1}
\label{alg:ext1}
\begin{algorithmic}[1]
\Require $x_1, \ldots, x_M$
\Ensure $P(\cdot|\allx)$
\State $\mathbf{p}_1 \gets \sigma\big(\cnn\!(x_1)\big)$
\For{$m \gets 2 \text{ to } M$}
    \State $\mathbf{p}_m \gets \sigma\big(\wsl_m\!(x_m)\big)$
\EndFor
\State $P(\cdot|\allx) \gets \frac{1}{M} \sum_{m=1}^M \mathbf{p}_m$
\end{algorithmic}
\end{algorithm}

\mypardot{\Ext{2}}
We propose to combine the sources in the logit level in this model, by taking a weighted sum of the logit vectors obtained from the reference source via reference CNN ($\cnn$) and the additional sources via weakly supervised instance attention networks ($\wsl_m$) using the following formulation:
\begin{equation}
    \comb\!(\allx) = \alpha_1 \cnn\!(x_1)
                       + \sum_{m=2}^{M} \alpha_m S^{-1}\!\big(\wsl_m\!(x_m)\big)
\label{eq:logitcomb}
\end{equation}
\begin{equation}
    P(c|\allx) = \big[\sigma\big(\comb\!(\allx)\big)\big]_c
\label{eq:condprob}
\end{equation}
where $S^{-1}$ is the inverse sigmoid function that maps WSL logits from the interval $[0, 1]$ to $(-\infty, \infty)$ to make them comparable to the logits obtained from the reference network. Weights $\alpha_m$ of the summation in \eqref{eq:logitcomb} are obtained using softmax over learnable parameters $\beta_m$: %
\begin{equation}
\label{eq:weightbeta}
    \alpha_m = \dfrac{\exp(\beta_m)}{\sum_{i=1}^{M} \exp(\beta_i)}.
\end{equation}
\rev{The \Ext{2} approach is summarized in Algorithm~\ref{alg:ext2}.}
In this formulation, since the sources with equally confident individual predictions can have different logits, the impact of each source on the final decision can be different. Conversely, even when a source has less confidence in a particular class than some other source, it could contribute more to the score of that class if the corresponding logit is larger. Therefore, combining the sources in the logit-level instead of probability-level aims to add more flexibility to the model in terms of each source's effect on the joint classification result.

\begin{algorithm}[t]
\caption{\Ext2}
\label{alg:ext2}
\begin{algorithmic}[1]
\Require $x_1, \ldots, x_M$
\Ensure $P(\cdot|\allx)$
\State $\mathbf{y}_1 \gets \cnn\!(x_1)$
\For{$m \gets 2 \text{ to } M$}
    \State $\mathbf{y}_m \gets S^{-1}\!\big(\wsl_m\!(x_m)\big)$
\EndFor
\State $P(\cdot|\allx) \gets \sigma\big(\sum_{m=1}^M \alpha_m \mathbf{y}_m\big)$
\end{algorithmic}
\end{algorithm}

\mypardot{\Ext{3}}
For each additional source $m$, we propose to combine penultimate layer feature vector of the reference network $\enc_\mathit{ref}(x_1)$ with the candidate region feature representations of each additional source $\enc_m(x_m)$. For this purpose, we replicate $\enc_\mathit{ref}(x_1)$ $R_m$ times, and concatenate with $\enc_m(x_m)$ to obtain fused feature vectors of size $F_\mathit{ref} + F_m$ for each of the $R_m$ candidate regions. The resultant vectors are processed by $\pred_m$ in the same way as the single-source model to obtain a logit vector per additional source. Finally, these logits are combined in the form of a weighted sum:
\begin{equation}
\label{eq:ext3}
    \comb\!(\allx) = \sum_{m=2}^{M} \alpha_m \pred_m\!\big(\psi(\enc_\mathit{ref}\!(x_1), \enc_m\!(x_m))\big)
\end{equation}
where $\psi$ denotes the aforementioned replication and concatenation operations. Class probabilities are obtained using \eqref{eq:condprob}.
Instead of an image-level combination, this approach focuses on utilizing the reference source earlier in the candidate region level. The idea behind this is to allow the model to leverage the lower-level information in the reference features and the candidate region features towards better classification and localization of the objects.
\rev{Algorithm~\ref{alg:ext3} provides a procedural description of \Ext{3}.}

\begin{algorithm}[t]
\caption{\Ext3}
\label{alg:ext3}
\begin{algorithmic}[1]
\Require $x_1, \ldots, x_M$
\Ensure $P(\cdot|\allx)$
\For{$m \gets 2 \text{ to } M$}
    \State $\mathbf{y}_m \gets \pred_m\!\big(\psi(\enc_\mathit{ref}\!(x_1), \enc_m\!(x_m))\big)$
\EndFor
\State $P(\cdot|\allx) \gets \sigma\big(\sum_{m=2}^M \alpha_m \mathbf{y}_m\big)$
\end{algorithmic}
\end{algorithm}

\mypardot{\Ext{4}}
Finally, we propose another form of a concatenation of the penultimate reference features with the additional sources. This time, instead of concatenating the reference feature vector with the feature vectors of the candidate regions obtained via $\enc_m$, we replicate and concatenate them directly to the pixels of the images of the additional sources ($x_m$), similar to the fusion technique in \citep{Sumbul:2019}. The fused input for the $m$\textsuperscript{th} source is then processed by $\wsl_m$ to obtain per-source logits. Finally, we take a weighted sum of the logits to obtain the combined logit vector:
\begin{equation}
    \comb\!(\allx) = \sum_{m=2}^{M} \alpha_m \wsl_m\!\big(\psi(\enc_\mathit{ref}\!(x_1), x_m)\big),
\end{equation}
which is followed by \eqref{eq:condprob} to obtain class probabilities. In this scheme, the motivation behind combining reference features with the input pixels is that a higher-level descriptor of the target object coming from the reference source could be useful in the pixel-level to guide the network towards a better localization, and therefore a better classification, of the object.
\rev{The \Ext{4} approach is summarized in Algorithm~\ref{alg:ext4}.}

\begin{algorithm}[t]
\caption{\Ext4}
\label{alg:ext4}
\begin{algorithmic}[1]
\Require $x_1, \ldots, x_M$
\Ensure $P(\cdot|\allx)$
\For{$m \gets 2 \text{ to } M$}
    \State $\mathbf{y}_m \gets \wsl_m\!\big(\psi(\enc_\mathit{ref}\!(x_1), x_m)\big)$
\EndFor
\State $P(\cdot|\allx) \gets \sigma\big(\sum_{m=2}^M \alpha_m \mathbf{y}_m\big)$
\end{algorithmic}
\end{algorithm}

\section{Experiments}
\label{sec:Experiments}

\newcommand{\PLH}{{\mkern-2mu\times\mkern-2mu}}

In this section, we first describe our experimental setup and implementation details for all methods. Then, we present our multisource results and compare them with other multisource methods as well as our single-source results.

\subsection{Experimental setup}
\label{sec:expsetup}

We conduct all experiments using two different multisource settings: (i) RGB \& MS, and (ii) RGB, MS \& LiDAR. The exact training procedure of each model differs from each other, especially in how they are pre-trained, which we observed to be very important on the final performance of the model. Here, we first outline the common aspects of the overall training procedure which is shared among all models. Then, we give the model-specific details about certain changes in the training procedure and hyper-parameters.

For all experiments, we randomly split the data set into training (60\%), validation (20\%), and test (20\%) sets. All of the models are trained on the training set using Adam optimizer with learning rate $10^{-3}$.  $\ell_2$-regularization with weight $10^{-5}$ is applied to all trainable parameters. These settings are same as in \citep{Sumbul:2019}. We use batches of size $100$ in each iteration. Each batch is drawn from an oversampled version of the training set to cope with the class imbalance. Oversampling rate for each class is proportional to the inverse frequency of that class. We augment the training set by shifting each image in both spatial dimensions with the amount of shift in each dimension randomly chosen between 0 and 20\% of the width/height. We adopt early-stopping with a patience value of $200$ to schedule the learning rate and terminate the training. If the validation accuracy does not improve for $200$ consecutive epochs, we first load the checkpoint with the highest accuracy and decrease the learning rate by a factor of $10$. If no improvement is observed for another $200$ epochs, we stop the training and choose the checkpoint with the highest validation accuracy for testing.
We use normalized accuracy as the performance metric where the per-class accuracy ratios are averaged to avoid biases towards classes with more examples.

\subsection{Implementation details}
\label{subsec:ImplementationDetails}

\mypardot{Single-source baseline classification networks}
We train three separate single-source classification networks for RGB, MS, and LiDAR. The basic network architectures (CNN) are taken from \citep{Sumbul:2019}. Dropout regularization is applied with a drop probability of $0.25$ in the convolutional layers and $0.5$ in the first fully-connected layer. We use the pre-trained RGB network to initialize the reference branch of all proposed multisource models and the pre-trained MS/LiDAR networks to initialize the MS/LiDAR branches. Such a pre-training strategy increases the validation score, which in turn improves the performance of the multisource models that are fine-tuned after being initialized.

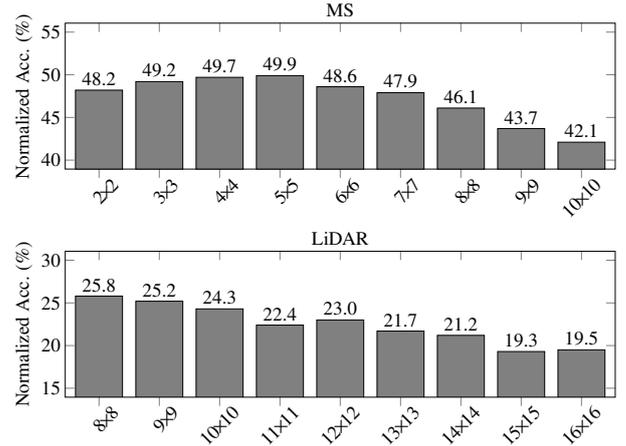
\begin{figure}
\centering
\begin{tikzpicture}[scale=0.8,transform shape,font=\small,inner sep=2pt]
\begin{axis}[
    ybar,
    bar width=22pt,
    height=4cm,
    width=1.2\linewidth,
    title=MS,
    title style={yshift=-5pt},
    ylabel near ticks,
    ylabel=Normalized Acc. (\%),
    ylabel shift=0pt,
    symbolic x coords={$2\PLH2$,
                       $3\PLH3$,
                       $4\PLH4$,
                       $5\PLH5$,
                       $6\PLH6$,
                       $7\PLH7$,
                       $8\PLH8$,
                       $9\PLH9$,
                       $10\PLH10$,},
    xticklabel style={rotate=45,anchor=east,xshift=5pt,yshift=-8pt},
    xtick=data,
    ytick={40,45,50,55},
    xtick pos=both,
    xtick align=center,
    xtick style={yshift=-2pt},
    ytick pos=both,
    ytick align=inside,
    ymin=40,
    ymax=55,
    enlargelimits=0.07,
    nodes near coords,
    nodes near coords style={/pgf/number format/.cd,fixed zerofill,precision=1},
]
\addplot [draw=black, fill=gray]
    coordinates {($2\PLH2$, 48.2)
                 ($3\PLH3$, 49.2)
                 ($4\PLH4$, 49.7)
                 ($5\PLH5$, 49.9)
                 ($6\PLH6$, 48.6)
                 ($7\PLH7$, 47.9)
                 ($8\PLH8$, 46.1)
                 ($9\PLH9$, 43.7)
                 ($10\PLH10$, 42.1)};
\end{axis}
\end{tikzpicture}
\begin{tikzpicture}[scale=0.8,transform shape,font=\small,inner sep=2pt]
\begin{axis}[
    ybar,
    bar width=22pt,
    height=4cm,
    width=1.2\linewidth,
    title=LiDAR,
    title style={yshift=-5pt},
    ylabel near ticks,
    ylabel=Normalized Acc. (\%),
    ylabel shift=0pt,
    symbolic x coords={$8\PLH8$,
                       $9\PLH9$,
                       $10\PLH10$,
                       $11\PLH11$,
                       $12\PLH12$,
                       $13\PLH13$,
                       $14\PLH14$,
                       $15\PLH15$,
                       $16\PLH16$,},
    xticklabel style={rotate=45,anchor=east,xshift=5pt,yshift=-8pt},
    xtick=data,
    ytick={15,20,25,30},
    xtick pos=both,
    xtick align=center,
    xtick style={yshift=-2pt},
    ytick pos=both,
    ytick align=inside,
    ymin=15,
    ymax=30,
    enlargelimits=0.07,
    nodes near coords,
    nodes near coords style={/pgf/number format/.cd,fixed zerofill,precision=1},
]
\addplot [draw=black, fill=gray]
    coordinates {($8\PLH8$, 25.8)
                 ($9\PLH9$, 25.2)
                 ($10\PLH10$, 24.3)
                 ($11\PLH11$, 22.4)
                 ($12\PLH12$, 23.0)
                 ($13\PLH13$, 21.7)
                 ($14\PLH14$, 21.2)
                 ($15\PLH15$, 19.3)
                 ($16\PLH16$, 19.5)};
\end{axis}
\end{tikzpicture}
\caption{Impact of region proposal size ($W$) in terms of normalized validation accuracy for the single-source instance attention models. Proposals are extracted within $12 \times 12$ pixel neighborhoods for MS and $24 \times 24$ pixel neighborhoods for LiDAR as described in Section \ref{sec:DataSet}. The smallest proposal sizes for MS and LiDAR are $2 \times 2$ and $8 \times 8$, respectively, because no pooling is used in the feature encoding network of the former whereas three pooling operations are included in the convolutional layers for the latter as described in Figure \ref{fig:wsddn}.}
\label{fig:regionsize}
\end{figure}

\mypardot{Single-source instance attention models}
Fully-connected layers of classification and localization branches of weakly supervised instance attention networks are initialized randomly while convolutional layers are initialized from the corresponding pre-trained baseline single-source classification networks. Similar to the basic classification network, we apply dropout with $0.25$ drop probability in the convolutional layers and $0.5$ drop probability in the first fully-connected layer. We choose the region size parameter $W$ as $5$ pixels for MS and $8$ pixels for LiDAR, which yield the highest validation accuracies in the experiments summarized in Figure \ref{fig:regionsize}.

Due to the multiplication of softmax outputs of classification and localization branches, output logits lie in the interval $[0, 1]$ when bias is not taken into account. Applying the final softmax operation before loss calculation with such logits results in smooth class distributions. Our experiments confirm that sharpening these distributions by introducing a temperature parameter ($T$) improves the performance of the model. With the addition of temperature, the final softmax in \eqref{eq:outputwsl} becomes:
\begin{equation}
    P(c|x)=\big[\sigma\big(\wsl\!(x)/T\big)\big]_c.
\end{equation}
Using our preliminary results on the validation set, we fix $T$ to $1/60$ for both MS and LiDAR.

\mypardot{\Ext1}
We observe that fine-tuning the network consisting of a pre-trained basic RGB network and pre-trained instance attention models combined as \ext1 does not improve the validation score. Furthermore, random initialization instead of pre-training worsens the network performance. Upon this observation, although it is possible to train/fine-tune the whole model end-to-end, we decide not to apply any fine-tuning. We choose temperature parameters ($T_m$) on the validation set via grid search, resulting in $1/48$ for MS and $1/18$ for LiDAR for the fused model.

\mypardot{\Ext2}
We initialize the RGB network and instance attention models from pre-trained models as in \ext1. $\beta_m$ parameters in \eqref{eq:weightbeta} are chosen as $1$ for RGB and $2.5$ for MS branch in the RGB \& MS setting; $1$ for RGB, $2.5$ for MS, and $1.5$ for LiDAR branch in the RGB, MS \& LiDAR setting using the validation set. We also observe the temperature parameter to be useful in this case as well, and set it to $0.25$ for both MS and LiDAR branches. The whole network is trained in an end-to-end fashion using dropout with a drop probability of $0.25$ in the convolutional and $0.5$ in the first fully-connected layers of all branches.

\mypardot{\Ext3}
Even though it is possible to train both MS and LiDAR branches of the model jointly in all-sources setting, we obtain a higher validation accuracy when we combine separately trained RGB \& MS and RGB \& LiDAR models. After individual training of the MS and LiDAR branches, we choose the logit combination weights $\alpha_m$ in \eqref{eq:ext3} on the validation set as $0.74$ for MS and $0.26$ for LiDAR to obtain the combined RGB, MS \& LiDAR classification results. As an alternative, we have tried incorporating logit combination similar to \eqref{eq:weightbeta} but it performed worse.

For the training of RGB \& MS and RGB \& LiDAR models, we initialize the whole RGB network from the pre-trained basic CNN model following the same approach as the previous models. For MS and LiDAR branches, convolutional layers are initialized from pre-trained instance attention models while fully-connected layers are initialized randomly, since the sizes of the fully-connected layers in classification and localization branches change due to feature concatenation. Furthermore, we observe that freezing all pre-trained parameters and training the rest of the models yields better validation accuracies. Although we freeze some of the network parameters, we find that leaving the dropout regularization on for the frozen layers improves the performance. For the RGB \& MS setting, we use a $0.5$ and $0.1$ drop probability for the convolutional and penultimate fully-connected layers, respectively, and $T$ is tuned to $0.05$. For the RGB \& LiDAR setting, we use a $0.1$ and $0.5$ drop probability for the convolutional and penultimate fully-connected layers, respectively, and $T$ is tuned to $0.025$.

\mypardot{\Ext4}
We make the same observation in this model as in \ext3 that combining separately trained RGB \& MS and RGB \& LiDAR models results in better validation performance than training both branches jointly. Similarly, we obtain higher validation accuracy for logit combination weights $\alpha_m$ chosen as $0.76$ for MS and $0.24$ for LiDAR using a grid search.

For the training of RGB \& MS and RGB \& LiDAR branches, the basic RGB network is initialized using the pre-trained model. Since the size of the first convolutional layer of the instance attention model is different from its single-source version, the first layer is initialized randomly. We also observe that random initialization of the classification and localization branches results in higher scores. Other layers are initialized from the pre-trained instance attention model and the whole network is fine-tuned end-to-end. The drop probability of dropout is chosen as $0.25$ for the convolutional layers and $0.5$ for the fully-connected layers. The temperature parameter is set to $1/60$ and kept constant as in the other models.

\subsection{Results}

\mypar{Single-source results and ablation study.}
We first evaluate the effectiveness of the instance attention framework in the case of single-source object recognition.
For this purpose, we compare the MS-only and LiDAR-only instance attention models to the corresponding single-source
baselines \rev{described in Section \ref{subsec:ImplementationDetails}, as well as a single-source spatial transformer network (STN) based model. For the latter, we adapt the methodology of \citet{he2019_stn} to our case and use an STN to select a candidate region from each input image. In the STN baseline, selected candidate regions are scaled to the same size as the input images and are classified by a CNN with the same architecture as the single-source baseline models. We restrict STN to use only translation and scaling transformations and use the same scale for both spatial dimensions, which results in three parameters to estimate. We estimate these parameters using a separate CNN with the same architecture as the single-source baseline model, but replace the final layer with a $3$-dimensional fully-connected layer}. We note that RGB is the reference high-resolution source and contains centered object instances,
therefore, instance attention \rev{and STN are} not applicable to RGB inputs.

\rev{Single-source results are presented in Table~\ref{tab:ablation}.} From the results we can see that instance attention
significantly improves both the MS-only results (from $40.6\%$ to $48.3\%$) and the LiDAR-only results ($21.2\%$ to $25.3\%$). The
essential reason for the large performance gap is the fact that single-source baselines aim to model the images holistically. This can be interpreted as separately modeling each potential instance location of each class when applied to a larger area, which is clearly very ineffective. In contrast, instance attention models rely on local recognition of image regions and attention-driven accumulation of local recognition results, which is much more resilient to positional ambiguity.
\rev{We also observe that instance attention yields consistently better results in comparison to STN on both MS and LiDAR inputs.}

As an ablative experiment, we additionally evaluate the importance of the localization branch (i.e., the $\zd$ component) in instance attention models. We observe that the localization branch improves the MS-only result from $47.7\%$ to $48.3\%$ and the LiDAR-only result from $24.3\%$ to $25.3\%$. These results show that the model with only the classification branch already performs significantly better than single-source baseline models thanks to handling object recognition locally. Incorporation of the localization branch further improves the results thanks to better handling of positional ambiguity.

Finally, we compare the MS-only and LiDAR-only instance attention models against the RGB-only single-source baseline. We observe that all MS models significantly outperform the RGB-only result, highlighting the value of detailed spectral information. We also observe that LiDAR is much less informative compared to MS, and, only the full single-source instance attention model for LiDAR is able to match the results of the RGB-only baseline model.

\begin{table}
\centering
\caption{Single-source baseline networks, single-source instance attention models, and ablation study results in terms of normalized test accuracy (\%).}
\label{tab:ablation}
\setlength{\tabcolsep}{2pt}
\begin{tabular}{lc}
\hline
Model & Accuracy\\
\hline
Single-source baseline (RGB) & 25.3 \\
\hline
Single-source baseline (MS) & 40.6 \\
\rev{Single-source STN \citep{he2019_stn} (MS)} & 41.1 \\
Instance attention, $\zc$ only (MS) & 47.7 \\
Instance attention (MS) & 48.3 \\
\hline
Single-source baseline (LiDAR) & 21.2 \\
\rev{Single-source STN \citep{he2019_stn} (LiDAR)} & 20.8 \\
Instance attention, $\zc$ only (LiDAR) & 24.3 \\
Instance attention (LiDAR) & 25.3 \\
\hline
\end{tabular}
\end{table}

\begin{figure}
\centering
\begin{tikzpicture}[scale=0.8,transform shape,font=\small,inner sep=2pt]
\begin{axis}[
    ybar,
    bar width=20pt,
    height=4cm,
    width=1.2\linewidth,
    ylabel near ticks,
    ylabel=Normalized Acc. (\%),
    ylabel shift=0pt,
    symbolic x coords={$8\PLH8$,
                       $12\PLH12$,
                       $16\PLH16$,
                       $20\PLH20$,
                       $24\PLH24$,
                       $28\PLH28$,
                       $32\PLH32$,
                       $36\PLH36$,
                       $40\PLH40$,
                       $44\PLH44$,
                       $48\PLH48$,},
    xticklabel style={rotate=45,anchor=east,xshift=5pt,yshift=-8pt},
    xtick=data,
    ytick={35,40,45,50,55,60},
    xtick pos=both,
    xtick align=center,
    xtick style={yshift=-2pt},
    ytick pos=both,
    ytick align=inside,
    ymin=35,
    ymax=60,
    enlargelimits=0.07,
    nodes near coords,
    nodes near coords style={/pgf/number format/.cd,fixed zerofill,precision=1},
]
\addplot [draw=black, fill=gray]
    coordinates {($8\PLH8$, 38.0)
                 ($12\PLH12$, 48.3)
                 ($16\PLH16$, 54.5)
                 ($20\PLH20$, 56.0)
                 ($24\PLH24$, 56.9)
                 ($28\PLH28$, 57.4)
                 ($32\PLH32$, 56.8)
                 ($36\PLH36$, 57.0)
                 ($40\PLH40$, 56.3)
                 ($44\PLH44$, 55.4)
                 ($48\PLH48$, 55.1)};
\end{axis}
\end{tikzpicture}
\caption{\rev{Effect of neighborhood size ($N$) in terms of normalized test accuracy for the MS-only instance attention model. $5 \times 5$ pixel proposals are extracted within these image neighborhoods.}}
\label{fig:neighborhood}
\end{figure}
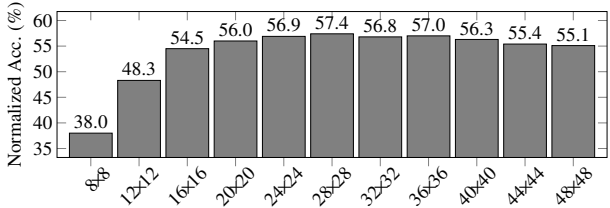

\mypar{Effect of neighborhood size.}
The proposed instance attention model uses $W \times W$ pixel windows as region
proposals within $N \times N$ neighborhoods as shown in Figure \ref{fig:wsddn}.
We have used $12 \times 12$ pixel neighborhoods for MS and $24 \times 24$ pixel
neighborhoods for LiDAR to be consistent with our previous work
\citep{Sumbul:2019}. To study the effect of different neighborhood sizes, we
perform additional experiments by fixing the region size parameter $W$ to the
best performing value of $5$ pixels as presented in Figure \ref{fig:regionsize}
and by varying the neighborhood size parameter $N$ for the MS data. The results
are presented in Figure \ref{fig:neighborhood}. We observe that the accuracy
increases beyond the previously used setting of $12 \times 12$ pixels until
the neighborhood size reaches $28 \times 28$ and starts to slightly decrease
afterwards. Even though this increase seems to imply the necessity of using
larger neighborhoods, it is important to note that there are other factors that
affect this performance. An analysis of the tree locations in the GIS data shows
that the average distance between neighboring trees in the MS data is slightly
above $6$ pixels. This means that increasing the window size too much also
increases the risk of including highly overlapping regions across training
and test samples. In addition, these relative accuracy improvements can also be
a superficial result of the exploitation of background patterns, which is more
likely to happen when the provided background contexts are large enough to be
informative for class discrimination.
In the rest of the experiments, therefore, we continue to use $12 \times 12$
pixels for MS and $24 \times 24$ pixels for LiDAR for the neighborhood size as
in \citet{Sumbul:2019}.

\begin{figure}
\centering
\begin{tikzpicture}[scale=0.85,transform shape,font=\small,inner sep=2pt]
\begin{axis}[
    width=1.2\linewidth,
    height=6cm,
    ylabel near ticks,
    ylabel=Normalized Acc. (\%),
    ylabel shift=0pt,
    xlabel=Number of Parameters,
    xtick align=center,
    ytick align=center,
    ymin=49, ymax=59,
    legend pos=south east,
    ymajorgrids=true,
    grid style=dashed,
    every axis plot/.append style={ultra thick},
]
\addplot[color=blue!40!gray,solid]
    coordinates {
    (714298,50.3)(5877040,56.8)(11114362,55.8)(15958903,56.1)(21698268,56.5)
    };
\addplot[color=red!40!gray,solid]
    coordinates {
    (714298,51.6)(5877040,56.6)(11114362,56.4)(15958903,56.2)(21698268,57.0)
    };
\addplot[color=green!40!gray,solid]
    coordinates {
    (719378,51.7)(5891840,58.0)(11134802,57.5)(15983423,57.1)(21726908,57.8)
    };
\addplot[color=yellow!40!gray,solid]
    coordinates {
    (782866,49.2)(6483214,54.8)(12273490,55.4)(17630785,55.5)(23979722,54.4)
    };
\legend{\Ext1,\Ext2,\Ext3,\Ext4}
\end{axis}
\end{tikzpicture}
\caption{Effect of width-wise increasing the number of parameters, hence the model capacity, in terms of normalized test accuracy for the RGB \& MS setting. This analysis aims to make a fair comparison among the multisource instance attention formulations.}
\label{fig:numparams}
\end{figure}
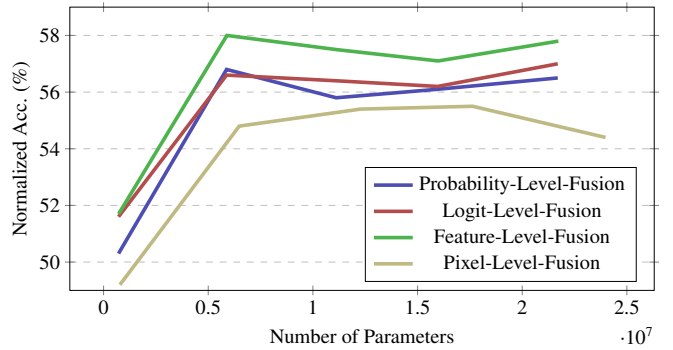

\mypar{Performance versus model capacity.}
We now examine how the proposed models perform with different model capacities for the RGB \& MS setting. We believe that it is immensely important to compare formulations with similar model complexities, and evaluate how their performances vary as a function of model complexity, to reach accurate conclusions. For that purpose, we use the number of parameters as a proxy for the model capacity, and train all models by keeping the network depth (i.e., number of layers) constant while increasing the network width (i.e., number of filters for the convolutional layers and number of output units for the fully-connected layers). We run these experiments in five different settings, in each of which the width is increased by a constant factor, starting from the default model capacity setting where the number of parameters of each method is comparable to the model in \citep{Sumbul:2019}.

Figure \ref{fig:numparams} shows the number of parameters of each model in each of these settings and their corresponding test scores. According to this, although \ext2 and \ext3 produce very similar results in the default setting with fewer parameters, the gap between \ext3 and the other methods increases as the model capacity increases, which points out that \ext3 is superior to other methods. Furthermore, all models' scores tend to increase up to some point before plateauing, except for \ext4, which starts to drop as the number of parameters increases, due to the 8-channel input size of MS being constant while the number of RGB features concatenated to them increasing to the point that they dominate the MS input.

The results of the model capacity experiments highlight two important points which are often overlooked when different models are compared in the literature. First, evaluating a model in a single capacity setting might yield sub-optimal results and prevent us from observing the full potential of the model. As an example, while \ext3, the best performing model according to Table \ref{tab:baselines}, achieves a test score of $51.7\%$ in the default setting, it shows a significantly higher performance of $58.0\%$ test accuracy with an increase in the model capacity. Second, comparing different methods in a single capacity setting might be an unreliable way of assessing the superiority of one method to another. For instance, the small difference of $0.1\%$ between the \ext2 and \ext3 scores in the default setting hinders us to reach a clear conclusion between the two methods. However, observation of a $1.4\%$ difference with a higher capacity enables us to verify \ext3's superiority. Furthermore, the performance difference between \ext2 and \ext1 closes or becomes reversed at different points as we increase the number of parameters.

\begin{table}
\centering
\caption{Multisource instance attention results and comparison to state-of-the-art in terms of normalized test accuracy (\%).}
\label{tab:baselines}
\def\myprefix{Instance attention - }
\setlength{\tabcolsep}{2pt}
\begin{tabular}{lc}
\hline
\multicolumn{2}{r}{Model \hfilll Accuracy}\\
\hline
\multicolumn{2}{l}{\textbf{Two sources (RGB \& MS)}}\\
Basic multisource model \citep{Sumbul:2019} & 39.1 \\
Recurrent attention model \citep{Fu:2017} & 41.6 \\
MRAN \citep{Sumbul:2019} & 46.6 \\
\myprefix \Ext1 & 50.3 \\
\myprefix \Ext2 & 51.6 \\
\myprefix \Ext3 & \textbf{51.7} \\
\myprefix \Ext4 & 49.2 \\
\hline
\multicolumn{2}{l}{\textbf{Three sources (RGB, MS \& LiDAR)}}\\
Basic multisource model \citep{Sumbul:2019} & 41.4 \\
Recurrent attention model \citep{Fu:2017} & 42.6 \\
MRAN \citep{Sumbul:2019} & 47.3 \\
\myprefix \Ext1 & 51.9 \\
\myprefix \Ext2 & 50.9 \\
\myprefix \Ext3 & \textbf{53.0} \\
\myprefix \Ext4 & 51.6 \\
\hline
\myprefix \Ext3 \emph{(inc.~capacity)} & 58.0 \\
\hline
\end{tabular}
\end{table}

\mypar{Comparison to the state-of-the-art.}
We compare the four proposed models against three state-of-the-art methods.
The first method is named the \textit{basic multisource model} that implements the commonly used scheme of extracting features independently from individual sources and concatenating them as the multisource representation that is used as input to fully-connected layers for the final classification. We use the end-to-end trained implementation in \citep{Sumbul:2019}.
The second method is the \textit{recurrent attention model} \citep{Fu:2017}. This model processes a given image at different scales using a number of classification networks. An attention proposal network is used to select regions to attend in a progressive manner. Classification networks are trained with intra-scale classification loss while inter-scale ranking loss, which enforces the next scale classification network to perform better than the previous scale, is used to train the attention proposal networks.
The third state-of-the-art method is the \textit{Multisource Region Attention Network} (MRAN)~\citep{Sumbul:2019}, which has been shown to be an effective method for multisource fine-grained object recognition. MRAN extracts candidate regions from MS and/or LiDAR data in a sliding window fashion and extracts features from these candidates by processing them with a CNN. The features are pooled in a weighted manner to obtain an attention-based representation for the corresponding source. Attention weights are obtained through a separate network that takes pixel-wise concatenation of RGB features, coming from the same basic single-source network architecture that we use, to the candidate regions as the input. The final multisource representation is obtained by the concatenation of RGB, MS and/or LiDAR representations, which is used for classification.

Table \ref{tab:baselines} lists the normalized test accuracies for the default model capacity setting (except for the bottom-most row), where the number of parameters is comparable to MRAN to enable comparisons with the state-of-the-art. Looking at these results, we see that all proposed methods outperform MRAN as well as the basic multisource model and the recurrent attention model. An interpretation for this could be that the instance attention is better suited to the classification task, arguably thanks to stronger emphasis on particular candidate regions. The last row of Table \ref{tab:baselines} shows the performance of \Ext3 for RGB \& MS in a higher model capacity setting with an $11.4\%$ improvement over MRAN for RGB \& MS, indicating that the model we propose can be scaled-up for increased performance. \Ext3 consistently outperforming \ext1 and \ext2 for all capacity settings in Figure \ref{fig:numparams} and both RGB \& MS and RGB, MS \& LiDAR settings in Table \ref{tab:baselines} indicates that combining the reference source with the additional sources earlier helps the network to better locate and classify the object of interest by making use of the additional information in the reference features which is not present in the logit level. The drop on the performance of \ext4, on the other hand, shows that fusing high-level reference features with low-level pixel values is not as effective as using reference features just before the classification and localization branches.

\begin{table}
\centering
\caption{Impact of data augmentation via random horizontal and vertical shifts in terms of normalized test accuracy (\%) for the RGB \& MS setting.}
\label{tab:augmentation}
\def\myprefix{Instance att. - }
\setlength{\tabcolsep}{2pt}
\begin{tabular}{lcc}
\hline
Model & \multicolumn{2}{c}{Accuracy}\\
\hline
& w/ aug. & w/o aug. \\
\myprefix \Ext1 & 50.3 & 49.7 \\
\myprefix \Ext2 & 51.6 & 50.1 \\
\myprefix \Ext3 & 51.7 & 50.4 \\
\myprefix \Ext4 & 49.2 & 47.5 \\
\hline
\end{tabular}
\end{table}

\mypar{Effect of data augmentation.}
As previously explained in Section \ref{sec:expsetup}, we use random shift based spatial data augmentation during training. In this part, we analyze the effect of this data augmentation policy on the recognition rates. For this purpose, we train and evaluate the multisource instance attention models for the RGB \& MS setting with and without data augmentation. The results presented in Table~\ref{tab:augmentation} show that data augmentation consistently improves each model by amounts varying from $0.6$ to $1.7$ points. We also observe that relative performances of the multisource models remain the same with and without data augmentation.

\begin{table}
\centering
\caption{Stability analysis of the multisource instance attention models in terms of normalized test accuracy (\%). Mean and standard deviation of the scores of five different runs are shown.}
\label{tab:stability}
\def\myprefix{Instance attention - }
\setlength{\tabcolsep}{2pt}
\begin{tabular}{lc}
\hline
\multicolumn{2}{r}{Model \hfilll Accuracy}\\
\hline
\multicolumn{2}{l}{\textbf{Two sources (RGB \& MS)}}\\
\myprefix \Ext1 \hspace{1em} & $50.5 \pm 0.8$ \\
\myprefix \Ext2 & $52.0 \pm 0.5$ \\
\myprefix \Ext3 & $51.5 \pm 0.8$ \\
\myprefix \Ext4 & $49.5 \pm 0.3$ \\
\hline
\multicolumn{2}{l}{\textbf{Three sources (RGB, MS \& LiDAR)}}\\
\myprefix \Ext1 & $51.9 \pm 0.4$ \\
\myprefix \Ext2 & $51.4 \pm 0.6$ \\
\myprefix \Ext3 & $53.1 \pm 0.6$ \\
\myprefix \Ext4 & $51.1 \pm 0.3$ \\
\hline
\end{tabular}
\end{table}

\mypar{Stability analysis.}
The previous experiments use a single split of the data set into training, validation, and test sets. In this part, we evaluate the stability of the multisource instance attention models under different partitionings of the data set. For this purpose, we use a random split of the data set into five folds. (All of the previous experiments correspond to the combination of the first three folds (60\%) as the training set, with the fourth fold (20\%) being the validation and the fifth fold (20\%) being the test sets.) For the stability analysis, we train and evaluate all models five times and report the mean and standard deviation of the test scores of these five runs. In each run, we use a unique combination of three folds as the training set, one of the remaining folds as the validation set, and the other fold as the test set. As a result, each of the five folds appears as an independent test set in the five runs. The results presented in Table~\ref{tab:stability} show that the performance variation across the folds is relatively small, which is reassuring about the stability of the models.

\begin{figure*}[t]
\centering
\includegraphics[width=\linewidth]{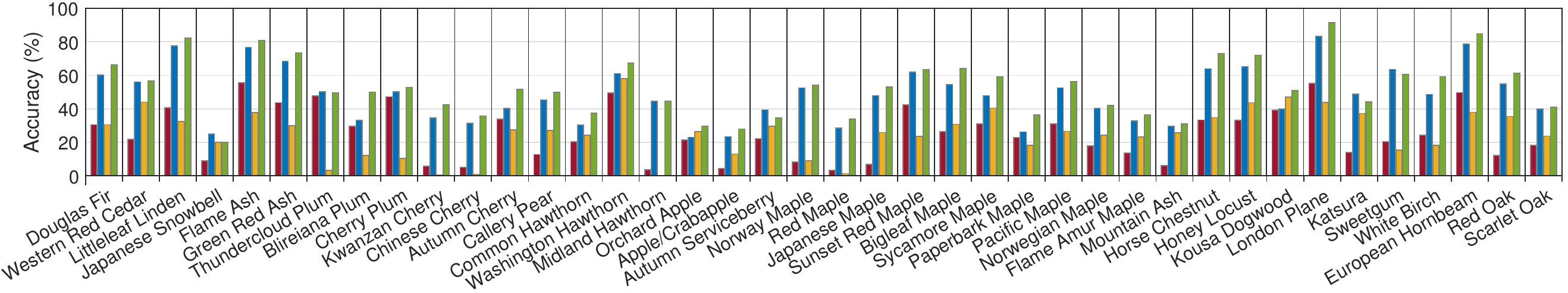}
\caption{Example results for the class-specific performances of the proposed methods. From left to right: single-source RGB (red), single-source MS with instance attention (blue), single-source LiDAR with instance attention (yellow), and \Ext3 (RGB, MS \& LiDAR) (green). Best viewed in color.}
\label{fig:ClassSpecific}
\end{figure*}

\begin{figure*}[t]
\centering
\includegraphics[width=\linewidth]{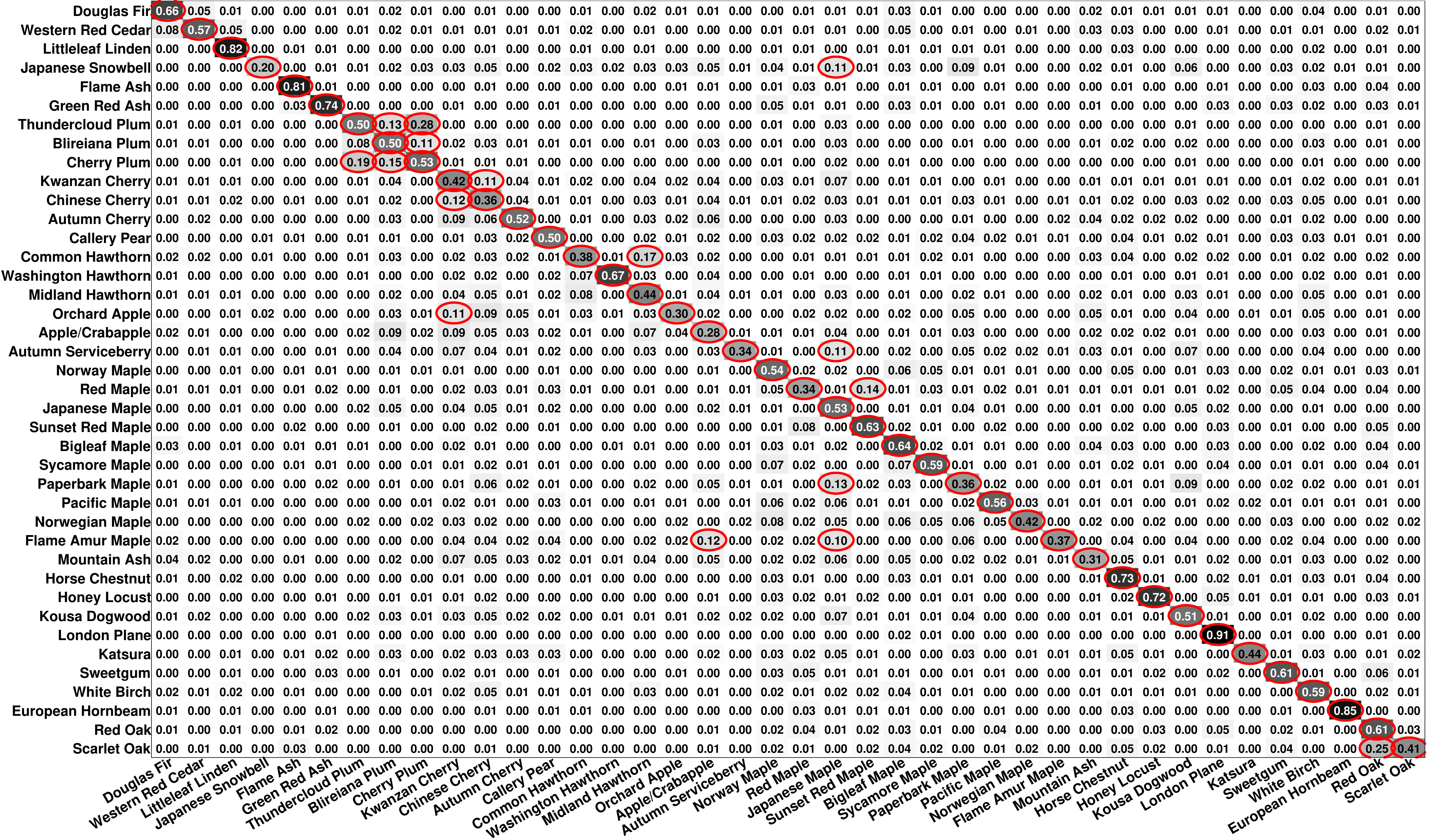}
\caption{Confusion matrix for \Ext3 (RGB, MS \& LiDAR). Values represent percentages where each row sums to $1$. The numbers greater than $0.10$ are marked with red ellipses. The ones on the diagonal show correct classification, whereas off-diagonal entries indicate notable confusions. The Kappa coefficient for this matrix is obtained as $0.5116$.}
\label{fig:ConfMatrix}
\end{figure*}

\mypar{Class-specific results.}
Figure \ref{fig:ClassSpecific} presents example results for the class-specific performances of the proposed methods. We observe that the classes receive different levels of contributions from different sources.
When we consider the models for the individual sources, the MS network performs significantly better than the RGB network where all classes have an improvement between $1\%$ and $44\%$ in accuracy. On the other hand, half of the classes have better performance under the RGB network compared to the other half that perform better with the LiDAR network.
When we compare the effect of the instance attention mechanism to the baseline single-source MS network, we observe that every one of the $40$ classes enjoys an improvement in the range from $1\%$ to $19\%$. Similarly, for the use of the attention mechanism in the single-source LiDAR network, $27$ of the classes receive higher scores with a maximum of $21\%$.
When we consider the best performing fusion model (\Ext3) under the RGB \& MS versus RGB, MS \& LiDAR settings, we observe that $30$ of the classes have improvements up to $7\%$ with the latter. Most of the classes that do not improve are among the ones with the least number of samples in the data set.
Finally, when the increased capacity network in the bottom-most row of Table \ref{tab:baselines} is compared to the default capacity one, the maximum improvement for the individual classes increases to $25\%$.
Overall, although the highest scoring model contains both MS and LiDAR sources, the contribution of the LiDAR data to the performance seems to be less significant compared to the MS data. This indicates that the richer spectral information in the MS images provides more useful information than the LiDAR data for the fine-grained classification task.
In addition, the proposed weakly supervised instance attention mechanism benefits the source (MS) with the smallest expected object size (maximum of $4 \times 4$ pixels) the most.

Figure \ref{fig:ConfMatrix} shows the confusion matrix resulting from the \Ext3 (RGB, MS \& LiDAR) model. We observe that most confusions are among the tree classes that belong to the same families in the scientific taxonomy shown in Figure \ref{fig:Taxonomy}. For example, $28\%$ of the thundercloud plum samples are wrongly predicted as cherry plum and $13\%$ are wrongly predicted as blireiana plum, whereas $19\%$ of the cherry plum samples are wrongly predicted as thundercloud plum and $15\%$ are wrongly predicted as blireiana plum. Similarly, Kwanzan cherry and Chinese cherry have the highest confusion with each other, with $11\%$ and $12\%$ misclassification, respectively, $17\%$ of common hawthorn are confused with midland hawthorn, and $25\%$ of scarlet oak are misclassified as red oak. As the largest family of trees, maples also have confusions among each other, with notable errors for red maple, paperback maple, and flame amur maple. In particular, flame amur maple has some of the highest confusions as being the class with the fewest number of samples.
As other examples for the cases with the highest confusion, $11\%$ of the Japanese snowbell samples and $11\%$ of autumn serviceberry are wrongly predicted as Japanese maple. All of these three types of trees have moderate crown density and have a spread in the $15\!-\!25$ feet range. Furthermore, autumn serviceberry and Japanese maple both have heights in the $15\!-\!20$ feet range (see \citep{Sumbul:2018} for a description of the attributes for the tree categories in the data set). As a final example, $11\%$ of orchard apple samples are wrongly predicted as Kwanzan cherry, with both species having moderate crown density, medium texture, and spread in the $15\!-\!25$ feet range.
Similar behaviors are observed for all other models. Since most of these types of trees are only distinguished with respect to their sub-species level in the taxonomy and have almost the same visual appearance, their differentiation using regions of few pixels from an aerial view is a highly challenging problem.
We think that the overall normalized accuracy of $53\%$ shows a significant performance for the fine-grained classification of $40$ different tree categories.

\newcommand{\mycorrect}{green}
\newcommand{\mywrong}{red}
\newcommand{\addfig}[6]{%
\begin{tikzpicture}[font=\footnotesize]
    \node[anchor=south west,inner sep=0] (image) at (0,0) {\includegraphics[width=0.5\linewidth]{#1}};
    \node[anchor=south west,inner sep=6,text=white] at (image.south west) {(#6)};
    \begin{scope}[x={(image.south east)},y={(image.north west)}]
        \draw[#2,solid,very thick] (0.431,0.066) rectangle (0.564,0.936);
        \draw[#3,solid,very thick] (0.571,0.066) rectangle (0.703,0.936);
        \draw[#4,solid,very thick] (0.710,0.066) rectangle (0.842,0.936);
        \draw[#5,solid,very thick] (0.848,0.066) rectangle (0.981,0.936);
    \end{scope}
\end{tikzpicture}
}

\begin{figure*}[t]
\centering
\addfig{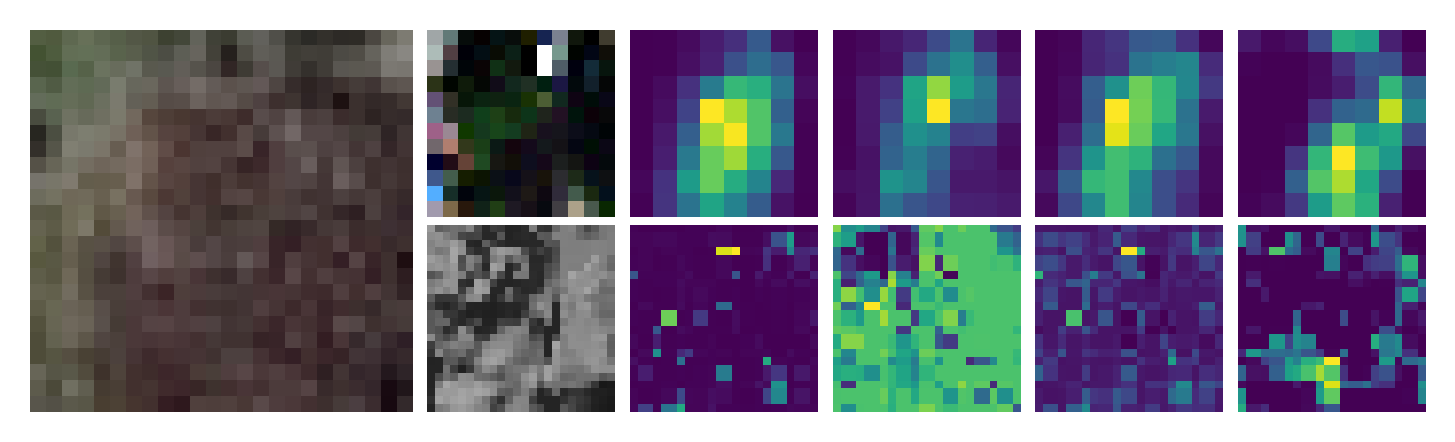}{\mycorrect}{\mycorrect}{\mycorrect}{\mycorrect}{a}\hspace{-8pt}
\addfig{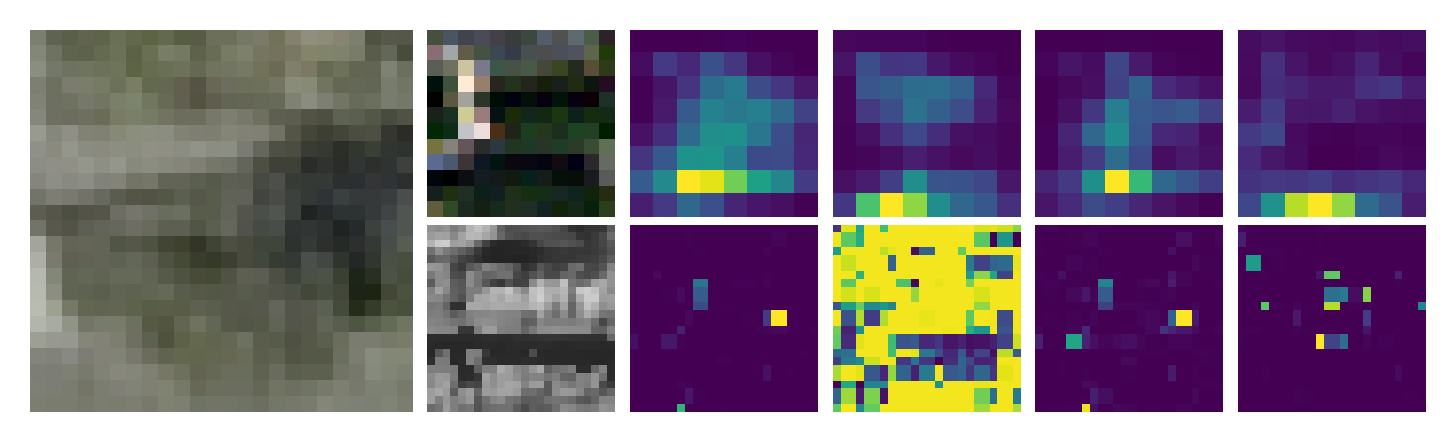}{\mycorrect}{\mywrong}{\mycorrect}{\mywrong}{b}\\[-4pt]
\addfig{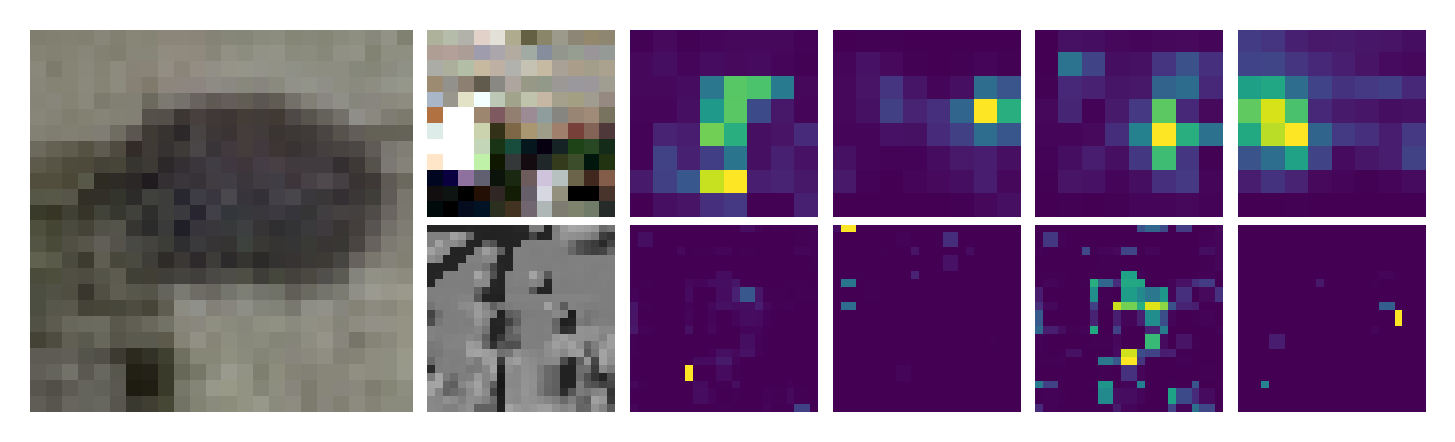}{\mywrong}{\mywrong}{\mycorrect}{\mywrong}{c}\hspace{-8pt}
\addfig{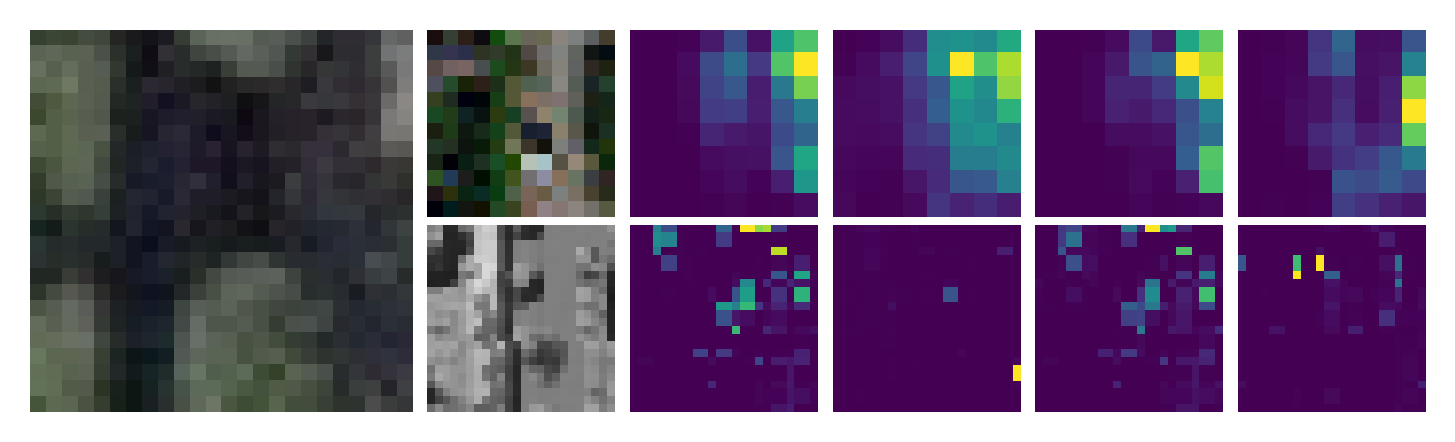}{\mywrong}{\mywrong}{\mywrong}{\mycorrect}{d}\\[-4pt]
\addfig{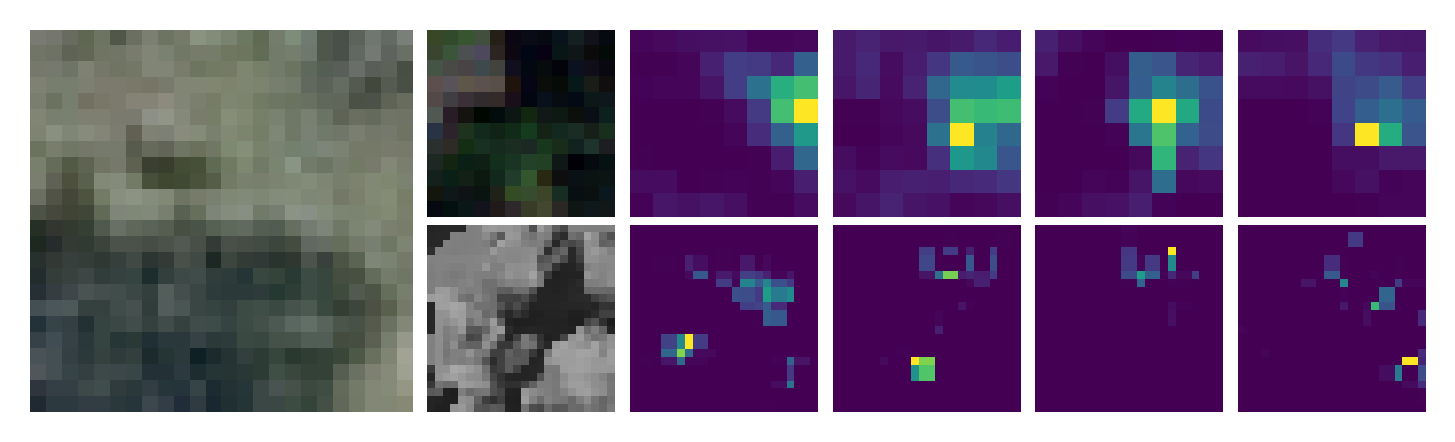}{\mycorrect}{\mywrong}{\mywrong}{\mycorrect}{e}\hspace{-8pt}
\addfig{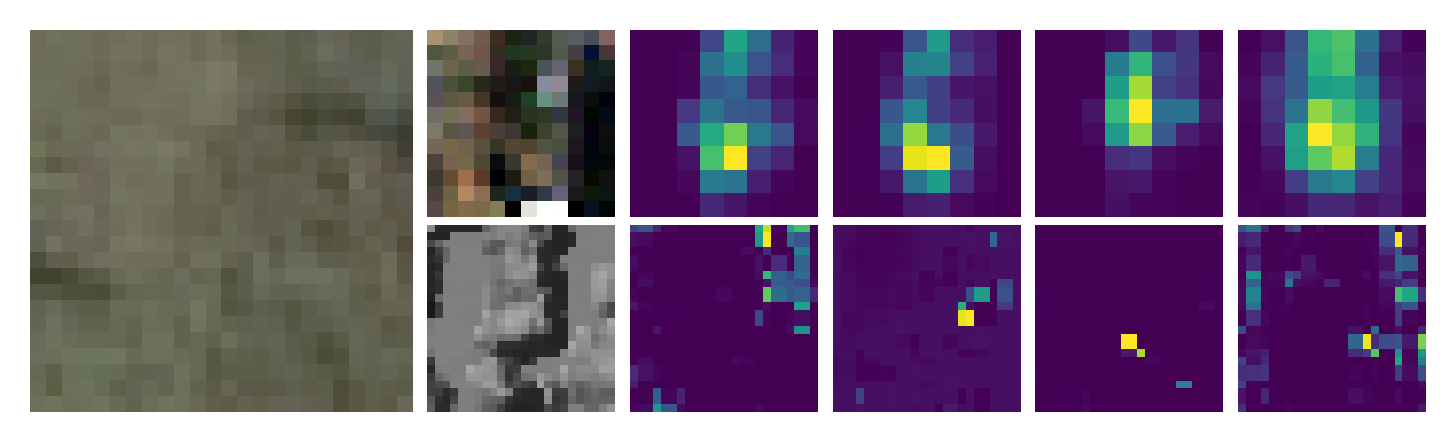}{\mycorrect}{\mycorrect}{\mywrong}{\mycorrect}{f}\\[-4pt]
\addfig{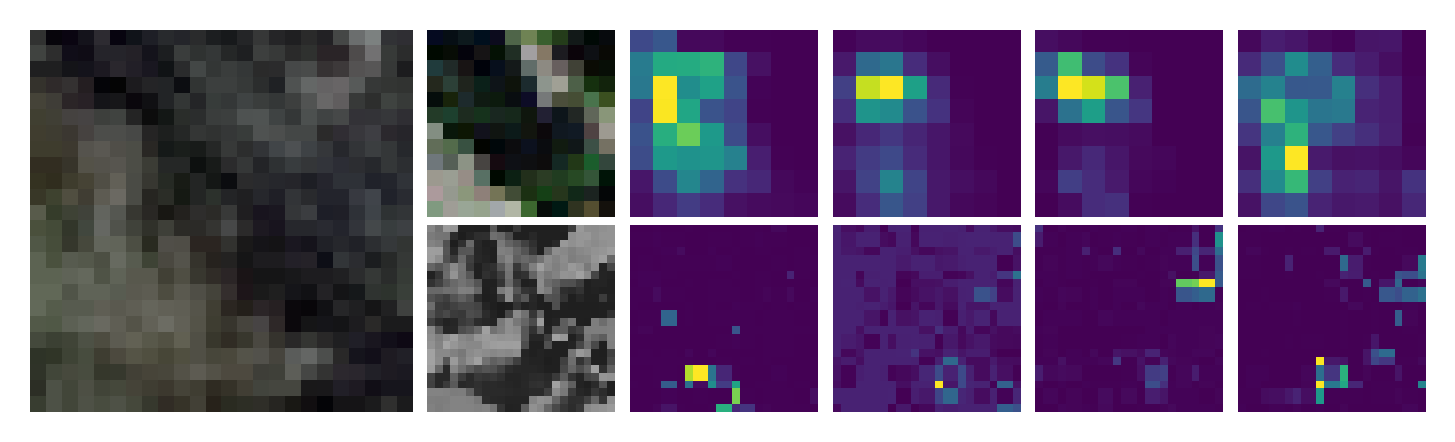}{\mywrong}{\mywrong}{\mywrong}{\mywrong}{g}\hspace{-8pt}
\addfig{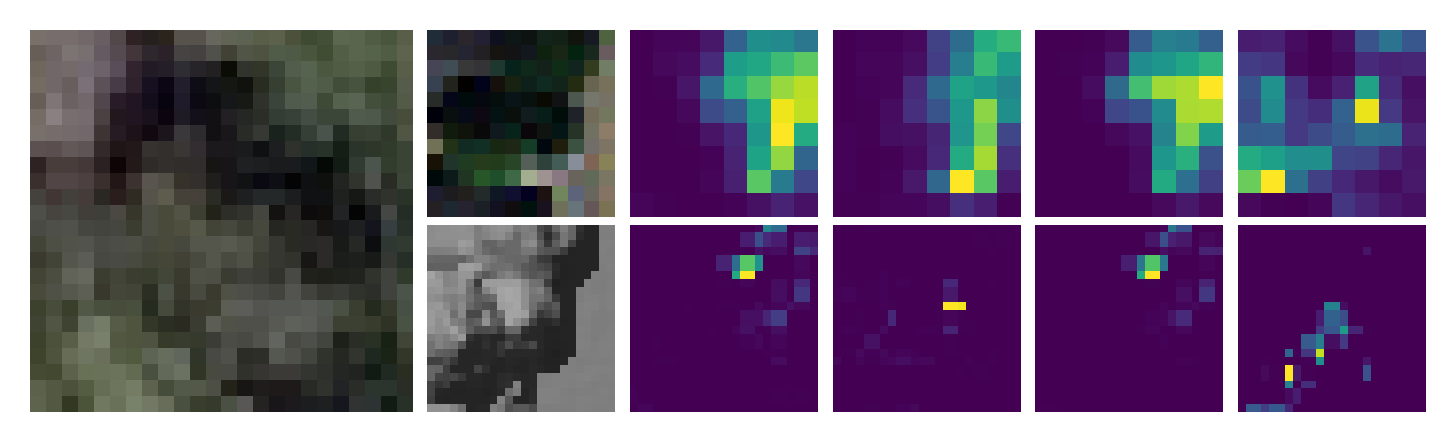}{\mycorrect}{\mycorrect}{\mycorrect}{\mywrong}{h}\\
\includegraphics[width=0.3\linewidth]{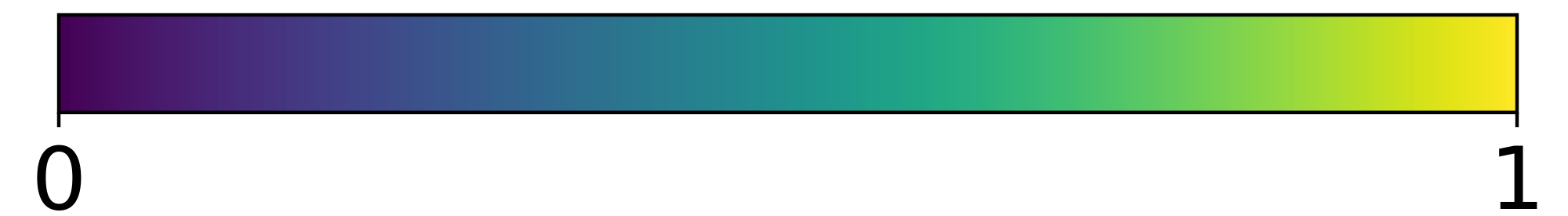}\\[-2pt]
\caption{Region scores for sample test images. RGB images are shown in the first column, MS (top) and LiDAR (bottom) neighborhoods shown in the second. Remaining columns show instance attention results respectively for \ext1, \ext2, \ext3, and \ext4. Results for correct class predictions are denoted with green boxes and those with wrong predictions are shown with red boxes. Region scores are obtained as the multiplication of per-region classification and localization scores corresponding to the predicted class. Best viewed in color.}
\label{fig:attentionVisualization}
\end{figure*}

\mypar{Qualitative results.}
Figure \ref{fig:attentionVisualization} illustrates the region scores, normalized to the $[0,1]$ range, obtained by multiplying per-region classification and localization scores in the Hadamard product in \eqref{eq:pred} for the predicted class.
\rev{Our first observation is that} the region scores for MS tend to have a smoother distribution with mostly a single local maximum, while LiDAR scores appear to be much noisier. This is in line with our previous observation that the information provided by the MS data appears to be more useful for the localization of the target object, which could explain its significantly higher contribution to the multisource classification results compared to LiDAR.

\rev{Figure \ref{fig:attentionVisualization}(a) supports this observation, where all methods successfully localize the objects in the MS image and classify the input correctly, even though \ext2 and \ext4 highlight different regions than the other methods for LiDAR. However, it is also possible to observe some cases as Figure \ref{fig:attentionVisualization}(b), where strong predictions by the LiDAR branch affect the final classification even when similar localization results are achieved by different methods for the MS data. In this particular case, the differences in the results of the LiDAR branches of \ext2 and \ext4 have an impact on the misclassification of the input.}

\rev{Next, we examine the localization results of the misclassified samples to better understand the effect of localization for the failure cases. For Figure \ref{fig:attentionVisualization}(c), only \ext3, which localizes the object differently than the other methods for both the MS and LiDAR data, is able to achieve a correct classification result. Looking at the corresponding MS image, we observe that the localization result indeed corresponds to a tree for \ext3, which points out that the misclassification of the other methods could be due to wrong localization. A similar case is seen in Figure \ref{fig:attentionVisualization}(d) as well, where only \ext4 succeeds at correctly classifying the input with a different localization than the others. However, even though MS and LiDAR inputs seem to coincide up to a certain degree for this particular example, the position of the localized object seems to differ a lot between the MS and LiDAR data, which highlights the possibility that even in the case of correct classification, the models can attend to contextual cues rather than the object itself.}

\rev{Even though localization has a substantial impact on the performance, we also observe failure cases for some samples such as Figure \ref{fig:attentionVisualization}(g), where the models output incorrect predictions even though the localization is successful. This result shows that we could achieve higher scores with the proposed approaches by improving their fine-grained classification performances in addition to their localization capabilities.}

\section{Conclusions}
\label{sec:Conclusions}

We studied the multisource fine-grained object recognition problem where the objects of interest in the input images have a high location uncertainty due to the registration errors and small sizes of the objects. We approached the location uncertainty problem from a weakly supervised instance attention perspective by cropping input images at a larger neighborhood around the ground truth location to make sure that an object with a given class label is present in the neighborhood even though the exact location is unknown. Using such a setting, we formulated the problem as the joint localization and classification of the relevant regions inside this larger neighborhood. We first outlined our weakly supervised instance attention model for the single-source setting. Then we provided four novel fusion schemes to extend this idea into a multisource scenario, where a reference source, assumed to contain no location uncertainty, can be used to help the additional sources with uncertainty to better localize and classify the objects.

Using normalized accuracy as the performance measure, we observed that all of the proposed multisource methods achieve higher classification scores than the state-of-the-art baselines with the best performing method (\ext3) showing a $5.1\%$ improvement over the best performing baseline using RGB \& MS data, and a $5.7\%$ improvement using RGB, MS \& LiDAR data. Additionally, we provided an in-depth comparison of the proposed methods with a novel evaluation scheme studying the effect of increased model capacity on the model performance. As a result of this experiment, we confirmed that \ext3 is indeed the most promising approach among all proposed methods, with an accuracy of $58.0\%$ using RGB \& MS data, which is a $6.3\%$ improvement compared to the default capacity setting.
Future work directions include the use of additional multisource fine-grained data sets for illustrating the generalizability of the proposed method, the use of additional measures such as the Kappa coefficient for performance evaluation, and the extension of the proposed model to handle other types of uncertainties such as temporal changes in addition to the spatial uncertainties studied in this paper.

\section*{Conflict of interest}

We confirm that there are no known conflicts of interest associated with
this work.

\section*{Acknowledgment}

This work was supported in part by the TUBITAK Grant 116E445 and in part by
the BAGEP Award of the Science Academy.

\bibliographystyle{elsarticle-harv}
\bibliography{definitions,bibliography}

\end{document}